\documentclass[10pt,journal,compsoc]{IEEEtran}
\usepackage{amsmath}
\usepackage{bbding}
\usepackage{amssymb}

\usepackage{epsfig}
\usepackage{graphicx}
\usepackage[utf8]{inputenc}
\usepackage[T1]{fontenc}
\usepackage{url}
\usepackage{nicefrac}
\usepackage{microtype}
\usepackage{lipsum}
\usepackage{wrapfig}
\usepackage{ragged2e}
\usepackage{booktabs}
\usepackage{amsfonts}
\usepackage{verbatim}
\usepackage{makecell}
\usepackage{multirow}
\newtheorem{definition}{Definition}
\newtheorem{theorem}{Theorem}

\newtheorem{remark}{Remark}
\newtheorem{proposition}{Proposition}
\newtheorem{problem}{Problem}

\usepackage{listings}
\usepackage{color}
\usepackage[dvipsnames]{xcolor}
\definecolor{dkgreen}{rgb}{0,0.6,0}
\definecolor{gray}{rgb}{0.5,0.5,0.5}
\definecolor{mauve}{rgb}{0.58,0,0.82}
\usepackage{bm}
\usepackage{caption}
\usepackage{booktabs,subcaption,amsfonts,dcolumn}
\usepackage[ruled,vlined]{algorithm2e}
\usepackage[switch]{lineno}
\usepackage[breaklinks=true,colorlinks,citecolor=blue,urlcolor=blue, bookmarks=false]{hyperref}
\newcommand{\setParDis}{\setlength {\parskip} {0.2cm} }
\newcommand{\setParDef}{\setlength {\parskip} {0pt} }
\newcommand{\bftab}{\fontseries{b}\selectfont}
\usepackage{fancyhdr}
\usepackage{colortbl}
\usepackage{pifont}
\definecolor{mygray}{gray}{.9}

\begin{document}
\title{Revisiting Confidence Estimation: Towards Reliable Failure Prediction}

\author{Fei Zhu,
	Xu-Yao Zhang,
	Zhen Cheng,
	%Zhaoxiang Zhang,
	Cheng-Lin Liu~\IEEEmembership{Fellow,~IEEE}
	\IEEEcompsocitemizethanks{
	\IEEEcompsocthanksitem This work was supported in part by the National Key Research and Development Program under Grant 2018AAA0100400, in part by the National Natural Science Foundation of China (NSFC) under Grants 61836014, 62222609, and 62076236, in part by the Key Research Program of Frontier Sciences of CAS (ZDBS-LY-7004), and in part by the CAS Project for Young Scientists in Basic Research (YSBR-083). (Corresponding author: Xu-Yao Zhang.)
	\IEEEcompsocthanksitem Fei Zhu is with the Centre for Artificial Intelligence and Robotics, Hong Kong Institute of Science and Innovation, Chinese Academy of Sciences, HongKong 999077, China.
	\IEEEcompsocthanksitem Xu-Yao Zhang, Zhen Cheng and Cheng-Lin Liu are with the State Key Laboratory of Multimodal Artificial Intelligence Systems, Institute of Automation of Chinese Academy of Sciences, 95 Zhongguancun East Road, Beijing 100190, P.R. China, and also with the School of Artificial Intelligence, University of Chinese Academy of Sciences, Beijing 100049, P.R. China.
	\IEEEcompsocthanksitem Email: \{zhufei2018, chengzhen2019\}@ia.ac.cn, \{xyz, liucl\}@nlpr.ia.ac.cn}
	
}

\IEEEtitleabstractindextext{
\begin{abstract}
%\justifying
   Reliable confidence estimation is a challenging yet fundamental requirement in many risk-sensitive applications. However, modern deep neural networks are often overconfident for their incorrect predictions, \emph{i.e.}, misclassified samples from known classes, and out-of-distribution (OOD) samples from unknown classes. In recent years, many confidence calibration and OOD detection methods have been developed. In this paper, we find a general, widely existing but actually-neglected phenomenon that most confidence estimation methods are harmful for detecting misclassification errors. 
   We investigate this problem and reveal that popular calibration and OOD detection methods often lead to worse confidence separation between correctly classified and misclassified examples, making it difficult to decide whether to trust a prediction or not. Finally, we propose to enlarge the confidence gap by finding flat minima, which yields state-of-the-art failure prediction performance under various settings including balanced, long-tailed, and covariate-shift classification scenarios. Our study not only provides a strong baseline for reliable confidence estimation but also acts as a bridge between understanding calibration, OOD detection, and failure prediction. The code is available at \url{https://github.com/Impression2805/FMFP}.
\end{abstract}

}

\maketitle
\IEEEdisplaynontitleabstractindextext
\IEEEpeerreviewmaketitle
\IEEEraisesectionheading{\section{Introduction}\label{sec:introduction}}

\pagestyle{fancy}
\cfoot{}
\rhead{\thepage}
\renewcommand{\headrulewidth}{0pt}
\renewcommand{\footrulewidth}{0pt}

\IEEEPARstart{D}{eep} neural networks (DNNs), especially vision models, have been widely deployed in safety-critical applications such as computer-aided medical diagnosis \cite{miotto2016deep, esteva2017dermatologist}, autonomous driving \cite{janai2017computer, bojarski2016end}, and robotics \cite{leidner2015classifying}. For such applications, besides the prediction accuracy, another crucial requirement is to provide \emph{reliable confidence} for users to make safe decisions. For example, an autonomous driving car should rely more on other sensors or trigger an alarm when the detection network is unable to confidently predict obstructions \cite{janai2017computer, zhu2023revisiting}. Another example is the control should be handed over to human doctors when the confidence of a disease diagnosis network is low \cite{miotto2016deep}. Unfortunately, modern DNNs are generally \emph{overconfident} for their false predictions, \emph{i.e.}, assign high confidence for misclassified samples from training classes and out-of-distribution (OOD) samples from unknown classes \cite{guo2017calibration, hendrycks2017baseline, havasi2020training, corbiere2019addressing}. The overconfident issue makes DNNs untrustworthy, bringing great concerns when deploying them in practical applications.

In recent years, many confidence estimation approaches have been proposed to enable DNNs to provide reliable confidence for their predictions. Most of those methods focus on two specific tasks, \emph{i.e.}, confidence calibration, and OOD detection. (1) \textbf{\emph{Confidence calibration}} alleviates the overconfidence problem and reflects the predictive uncertainty by matching the accuracy and confidence scores \cite{minderer2021revisiting}. One category of approaches \cite{pereyra2017regularizing, muller2019does, thulasidasan2019mixup, YunPLS20, XingAZP20, MukhotiKSGTD20, wen2020combining, zhong2021improving, Hebbalaguppe2022CVPR, Liu2022CVPR} aim to learn well-calibrated models during training. Another class of approaches \cite{guo2017calibration, RahimiSC0B20, KullFF17, ShehzadBFARKK20, gupta2020calibration, patel2020multi} use post-processing techniques to calibrate DNNs. (2) \textbf{\emph{OOD detection}} focuses on judging whether an input sample is from unseen classes based on model confidence. Most existing methods tackle OOD detection problem in a post-hoc manner \cite{LiangLS18, lee2018simple, liu2020energy, huang2021importance, sun2021react, hendrycks2019anomalyseg}, and other works focus on learning a model with better OOD detection ability at training time \cite{hendrycks2019deep, wei2022logitnorm, techapanurak2020hyperparameter, cheng2023unified}. If we focus on the progress made in \emph{confidence calibration} and \emph{OOD detection}, it seems that the confidence estimation of DNN is well-solved because both the confidences of in-distribution (InD) and OOD samples are well-calibrated.

\begin{figure*}[t]
	\begin{center}
		\centerline{\includegraphics[width=0.87\textwidth]{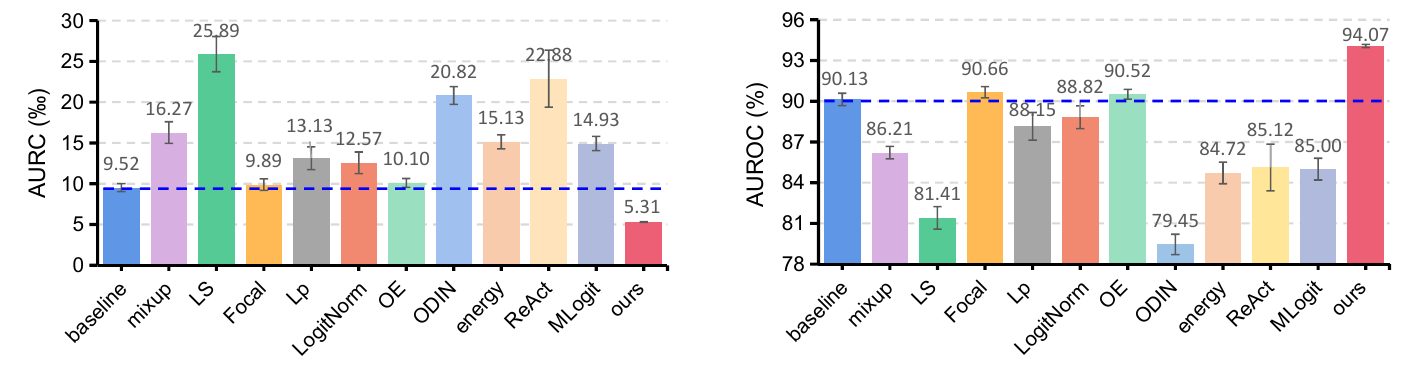}}
		\vskip -0.1 in
		\caption{A comparison of (a) AURC ($\downarrow$) and (b) AUROC ($\uparrow$). We observed that many popular confidence calibration and OOD detection methods are useless or harmful for failure prediction. We propose a simple flat minima based method that can yield state-of-the-art failure prediction performance. ResNet110 \cite{HeZRS16} on CIFAR-10 \cite{krizhevsky2009learning}.}
		\label{figure-1}
	\end{center}
	\vskip -0.25 in
\end{figure*}
In this paper, we study a natural but ignored question: can we use calibrated confidence to detect misclassified samples by filtering out low-confidence predictions? This, perhaps, is the most direct and practical way to evaluate the reliability of the estimated confidence. Actually, this problem is studied in the literature as \textbf{\emph{failure prediction}} (or misclassification detection) \cite{hendrycks2017baseline, zhu2023openmix, corbiere2021confidence}, whose purpose is to detect erroneously classified natural examples from seen class (\emph{e.g.}, misclassified samples in test set). In practice, for a deployed system, most of the inputs are from seen classes, and misclassification errors widely exist. Therefore, failure prediction is a highly practical and valuable tool for developing reliable and trustworthy machine learning systems \cite{corbiere2021confidence}. However, compared with the widely studied confidence calibration and OOD detection, there are few works for failure prediction. 

We revisit current confidence estimation approaches from the perspective of failure prediction motivated by the following reasons: \emph{\textbf{Firstly}}, calibration and failure prediction both focus on in-distribution data from known classes and share the same motivation that enables the model to provide reliable confidence to make safe decisions. From a practical perspective, with a calibrated classifier in hand, one natural way to verify its trustworthiness is to filter out predictions with low confidence. \emph{\textbf{Secondly}}, existing OOD detection methods aim to separate InD samples from OOD samples by assigning high confidence for InD samples and low confidence for OOD samples. However, in practice, misclassified samples should also have low confidence so that be separated from correct samples. In other words, OOD detection and failure prediction should be considered in a unified manner and a good confidence estimator should help detect both the OOD and misclassified InD samples. \emph{\textbf{Thirdly}}, calibration and OOD detection have drawn significant attention from the machine learning community \cite{muller2019does, thulasidasan2019mixup, YunPLS20, guo2017calibration, ovadia2019can, wen2020combining, minderer2021revisiting, LiangLS18, lee2018simple, liu2020energy, huang2021importance, sun2021react, hendrycks2019anomalyseg, hendrycks2019deep, wei2022logitnorm, techapanurak2020hyperparameter, yang2020convolutional}. However, there are few works for failure prediction, which is a practical, important, yet somewhat under-appreciated area of research that represents a natural testbed to evaluate the reliability of the estimated confidence of a given method. Revisiting confidence estimation methods from the perspective of failure prediction not only helps us understand the effectiveness of existing methods but also benefits the investigation of failure prediction.

Although common wisdom in the community suggests that the estimated confidence via calibration or OOD detection methods could be useful for failure prediction, we find a surprising pathology: many popular calibration and OOD detection methods (including both training-time \cite{pereyra2017regularizing, muller2019does, thulasidasan2019mixup, YunPLS20, HendrycksMCZGL20, hendrycks2019deep, wei2022logitnorm} and post-processing \cite{guo2017calibration, LiangLS18, liu2020energy, sun2021react, hendrycks2019anomalyseg} methods) are more of a hindrance than a help for failure prediction, as shown in Fig.~\ref{figure-1}. Empirical study shows that calibration methods often reduce overconfidence by simply aligning the accuracy and average confidence, and OOD detection methods lead to under-confident correctly classified InD samples. Consequently, correctly classified and misclassified samples would be worse separated, which is harmful for detecting misclassified samples based on the confidence of predictions.

Finally, how can we improve the failure prediction performance of DNNs? On the one hand, failure prediction requires better discrimination between the confidence of correctly classified and misclassified examples, which would increase the difficulty of changing the correctly classified samples to be incorrect due to the larger confidence margins. Interestingly, this is closely related to the notion of ``\emph{flatness}'' in DNNs, which reflects how sensitive the correctly classified examples become misclassified when perturbing model parameters \cite{huang2020understanding, izmailov2018averaging, foret2020sharpness, zhu2023imitating}. On the other hand, we observed an interesting \emph{reliable overfitting} phenomenon that the failure prediction performance can be easily overfitted during the training of a model. 
Inspired by the natural connection between flat minima and confidence separation as well as the effect of flat minima for overfitting mitigation, we propose a simple hypothesis: \emph{flat minima is beneficial for failure prediction}. We verify this hypothesis through extensive experiments and propose a simple flat minima based method that can achieve state-of-the-art confidence estimation performance. 

In summary, our contributions are as follows:
\begin{itemize}
	\item We rethink the confidence reliability of confidence calibration and OOD detection methods by evaluating them on the challenging and practical failure prediction task. Surprisingly, we find that they often have negative effect on failure prediction.
	\item We provide detailed analysis and discussion on calibration and OOD detection for failure prediction from the perspective of proper scoring rules and Bayes-optimal reject rules, respectively. 
	\item We reveal an interesting reliable overfitting phenomenon that the failure prediction performance can be easily overfitting. This phenomenon exists in different model and dataset settings.
	\item We propose to find flat minima to significantly reduce the confidence of misclassified samples while maintaining the confidence of correct samples. To this end, a simple flat minima based technique is proposed.
	\item Extensive experiments in balanced, long-tailed, and covariate-shift classification scenarios demonstrated that our method achieves state-of-the-art confidence estimation performance.
\end{itemize}

This paper extends our previous conference publication \cite{zhuccformd} mainly in five aspects: \textbf{(1)} We show novel insight into
the behavior of popular OOD detection methods for failure prediction based on comprehensive experiments. \textbf{(2)} We provide more theoretical perspectives based on proper scoring rules and Bayes-optimal reject rules for the failures of calibration and OOD detection methods in detecting misclassified instances. \textbf{(3)} We provide theoretical
analysis on the benefit of flat minima for improving confidence estimation based on PAC-Bayes framework. 
\textbf{(4)} We design more challenging and realistic benchmarks, \emph{i.e.}, confidence estimation under long-tailed and covariate-shift scenarios, in which our method also establishes state-of-the-art performance. \textbf{(5)} Experiments on standard OOD detection benchmarks demonstrate the strong OOD detection ability of our method.  In conclusion, we rethink the reliability of current calibration and OOD detection methods from the perspective of failure prediction. Our finding is important because it allows us to better assess recently reported progress in the area of confidence estimation. Finally, we provide a strong and unified baseline that can improve calibration and detect both misclassified and OOD samples. 

The remainder of this paper is organized as follows: Section 2 presents the problem formulation and background of confidence calibration, OOD detection and failure prediction, respectively. Section 3 evaluates and analyzes the effect of popular calibration, and OOD detection methods for failure prediction. Section 4 shows that failure prediction can be 
significantly improved by finding the flat minima in balanced, long-tailed and covariate-shift classification scenarios. Section 5 provides concluding remarks.

\section{Problem Formulation and Background}
\textbf{Multi-class classification.} Considering a $K$-class classification task, let $(X, Y) \in \mathcal{X} \times \mathcal{Y}$ be jointly distributed random variables, where $\mathcal{X} \subset \mathbb{R}^d$ denotes the in-distribution feature space and $\mathcal{Y}$ is the label space. Specifically, $\mathcal{Y} = \{e_1,...,e_K\}$ where $e_k \in \{0,1\}^{K}$ is the one-hot vector with 1 at the $k$-th index and 0 otherwise.
We are given labeled samples $\mathcal{D}=\{(\bm{x}_i, y_i)\}^n_{i=1}$ drawn \emph{i.i.d.} from $(X, Y)$ with density
$p(\bm{x}, y)$. We assume that labels are drawn according to the true posterior distribution $Q = (Q_1,..., Q_K) \in \Delta^K$ where $Q_k:=
\mathbb{P}(Y = e_k|X)$ and $\Delta^K$ is the probability simplex $\Delta^K = \{(p_1, . . . , p_K) \in [0,1]^K: \sum_{k}p_k=1 \}$.
A DNN classifier $f$ predicts the class of an input example $\bm{x}$:
\begin{equation}
\label{eq0}
\begin{aligned}
f(\bm{x}) &= \mathop{\arg\max}\limits_{k=1,...,K}p_k(\bm{x}), \\ p_k(\bm{x}) &= \text{exp}(\bm{z}_k(\bm{x}))/\sum_{k'=1}^{K}\text{exp}(\bm{z}_{k'}(\bm{x})),
\end{aligned}
\end{equation}
where $\bm{z}_k(\bm{x})$ is the logits output of the network with respect to class $k$, and $p_k(\bm{x})$ is the
probability (softmax on logits) of $\bm{x}$ belonging to class $k$.
The classification risk on classifier $f$ can be defined
with respect to the 0-1 loss as:
\begin{equation}
\label{eq1}
\mathcal{R}_{0\text{-}1}(f) = \mathbb{E}_{\substack{p(\bm{x}, y)}}[\mathbb{I}(f(\bm{x}) \neq y)],
\end{equation}
where $\mathbb{I}$ is the indicator function.
The Bayes-optimal classifier $f^{*}$ can be defined as follows:
\begin{definition} (\textbf{Bayes-optimal classifier})
	The Bayes-optimal solution of multi-class classification, $f^{*} = \mathop{\arg\min}\limits_{f}\mathcal{R}_{0\text{-}1}(f)$ can be expressed as $f^{*}(\bm{x}) = {\arg\max}_{y \in \mathcal{Y}}\mathbb{P}(y|\bm{x}).$
\end{definition} 
In practice, the risk in Eq.~(\ref{eq1}) is not straightforward to minimize because the number of training samples is finite and minimizing the zero-one loss
is known to be computationally infeasible. Therefore, we often minimize an empirical surrogate
risk \cite{bartlett2006convexity, vapnik1999nature}. Specifically, let $\ell$ be a surrogate loss (\emph{e.g.}, the log-loss), based on the empirical risk minimization approach \cite{vapnik1999nature}, the following empirical surrogate risk is minimized: 
$\mathcal{R}_{\ell}(f) = \frac{1}{n}\sum_{i=1}^{n}[\ell(f(\bm{x}),y)]$,
where regularization can be added to avoid overfitting.
\begin{figure*}[t]
	\begin{center}
		\centerline{\includegraphics[width=0.97\textwidth]{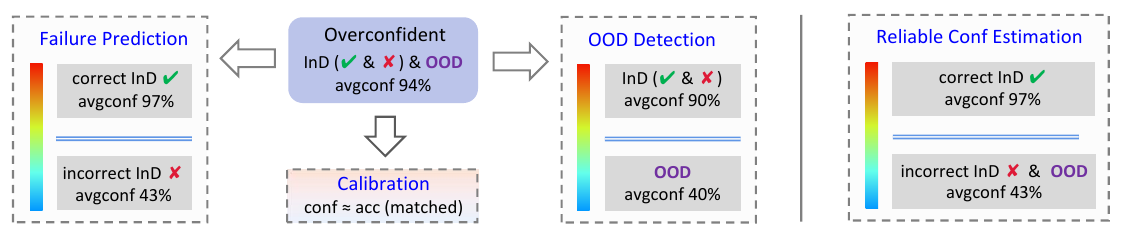}}
		\vskip -0.05 in
		\caption{Calibration reduces the mismatch between confidence and accuracy, and OOD detection distinguishes OOD samples from InD samples. They share the same motivation to provide reliable confidence for trustworthy AI. In practice, both OOD and misclassified samples are failure sources and should be rejected together.}
		\label{figure-2}
	\end{center}
	\vskip -0.2 in
\end{figure*}
At inference time, $\hat{y} = f(\bm{x})
$ can be returned as the predicted class and the associated probabilistic score $p_{\hat{y}}(\bm{x})$, \emph{i.e.}, the \emph{\textbf{maximum softmax probability (MSP)}}, can be viewed as the predicted confidence. Besides directly using MSP as confidence, other score functions (\emph{e.g.}, entropy, energy score \cite{liu2020energy} or maxlogit score \cite{hendrycks2019anomalyseg}) can also be used.

\subsection{Confidence Calibration}
Confidence calibration focuses on InD samples and aims to calibrate the confidence of a model to indicate the actual likelihood of correctness \cite{guo2017calibration}. For example, if a calibrated model predicts a set of inputs to be class $y$ with $40\%$ probability, then we expect $40\%$ of the inputs indeed belong to class $y$. Formally, denote the score $S = f(X)$ with $S = (S_1,...,S_K) \in \Delta^K$ the random vector describing the scores output by the probabilistic classifier $f$, then we state:
\begin{problem} \label{def1}
(\emph{\textbf{Calibrated}}).
A classifier giving scores $s = (s_1,..., s_k)$ is jointly calibrated if among instances getting score $s$, the class probabilities are equal to $s$:
\begin{equation}
\label{eq3}
\mathbb{P}(Y = e_k|S = s) = s_k ~~~~~\textit{for}~ k = 1,...,K.
\end{equation}
\end{problem}
In practice, it is challenging to estimate the probability of $Y$ conditioned on $S$ because the number of samples is limited \cite{perez2022beyond}. Therefore, in the machine learning community, the commonly used notion of calibration is a weaker definition \cite{guo2017calibration}: if among instances for which the confidence score of the predicted class is $s$, the probability that the prediction is correct is $s$: $\mathbb{P}(Y = e_{\mathop{\arg\max}(s)}|\mathop{\max}(S) = s) = s.$

The most commonly used calibration estimator is the Expected Calibration Error (ECE) \cite{Naeini2015ObtainingWC}, which approximates the miscalibration by binning the confidence in $[0,1]$ under $M$ equally-spaced intervals \emph{i.e.,} $\{B_m\}_{m=1}^M$. Then the miscalibration is estimated by taking the expectation of the mismatch between the accuracy and averaged confidence in each bin: $\text{ECE} = \sum_{m=1}^M \frac{\left|B_{m}\right|}{n} \left|\text{acc}(B_{m})-\text{conf}(B_{m})\right|$,
where $n$ is the number of all samples. Alternatives to ECE include the negative log-likelihood (NLL) and Brier score \cite{brier1950verification}.

\setParDis
\noindent
\textbf{Improving calibration.} Many strategies have been proposed to address the miscalibration of modern DNNs. (1) One category of approaches \cite{pereyra2017regularizing, muller2019does, thulasidasan2019mixup, YunPLS20, XingAZP20, MukhotiKSGTD20, wen2020combining, zhong2021improving} aim to learn well-calibrated models during training. For example, several works \cite{thulasidasan2019mixup, mixCalibration, howmixup} found that the predicted scores of DNNs trained with mixup \cite{zhang2018mixup} are better calibrated. Muller \emph{et al}. \cite{muller2019does} showed the favorable calibration effect of label smoothing. Mukhoti \emph{et~al}. \cite{MukhotiKSGTD20} demonstrated that focal loss \cite{LinGGHD20} can automatically learn well-calibrated models. 
(2) Another class of approaches \cite{guo2017calibration, RahimiSC0B20, KullFF17, ShehzadBFARKK20, gupta2020calibration, patel2020multi} rescale the predictions in a post-hoc manner. Among them, temperature scaling (TS) \cite{guo2017calibration} is an effective and simple technique, which has inspired various post-processing approaches \cite{KullPKFSF19, Mozafari2019UnsupervisedTS, ShehzadBFARKK20}. 

\noindent
\textbf{Empirical studies of calibration.}
In addition to the calibration strategies, there have been some empirical studies on calibration \cite{guo2017calibration, ovadia2019can, RethinkingCalibration, minderer2021revisiting}. For instance, Ovadia \emph{et al}. \cite{ovadia2019can} studied the calibration under distribution shift and empirically found the generally existing performance drop of different calibration methods under distribution shift. Minderer \emph{et al}. \cite{minderer2021revisiting} found that the most recent non-convolutional models \cite{dosovitskiy2020image, tolstikhin2021mlp} are well-calibrated, suggesting that architecture is an important factor of calibration.
\setParDef

\subsection{OOD Detection}
The goal of OOD detection is to reject OOD samples from InD samples at test time. Formally, we have an InD joint distribution $\mathcal{D}_{X_{\text{in}}Y_{\text{in}}}$, where $X_{\text{in}} \in \mathcal{X}$ and $Y_{\text{in}} \in \mathcal{Y}$ are random variables.  We also have an OOD joint
distribution $\mathcal{D}_{X_{\text{out}}Y_{\text{out}}}$, where $X_{\text{out}} \in \mathbb{R}^d\setminus\mathcal{X}$ and $Y_{\text{out}} \notin \mathcal{Y}$. At test time, we encounter a mixture of InD and OOD joint distributions: $\mathcal{D}_{XY} = \pi_{\text{in}} \mathcal{D}_{X_{\text{in}}Y_{\text{in}}} + (1-\pi_{\text{in}})\mathcal{D}_{X_{\text{out}}Y_{\text{out}}}$, and can only observe the marginal distribution $\mathcal{D}_{X} = \pi_{\text{in}} \mathcal{D}_{X_{\text{in}}} + (1-\pi_{\text{in}})\mathcal{D}_{X_{\text{out}}}$, where the $\pi_{\text{in}} \in (0, 1)$ is an unknown prior probability.

\begin{remark} Consistent with the mainstream of existing works \cite{hendrycks2017baseline, hendrycks2019anomalyseg, liu2020energy, bitterwolf2022breaking, zhu2022learning}, the OOD detection in our paper refers to detecting new-class (semantic) shifted examples that do not have overlapping labels w.r.t. training classes. In open environments, OOD examples could come from various domains, and can be far or near to InD. To make it possible to detect OOD examples, we assume that $X_{\text{out}}$ and $X_{\text{in}}$ are in different subsets of the input space $\mathbb{R}^d$. However, more generalized OOD detection \cite{yang2021generalized} might involve detecting covariate shifted examples, in which case the definition of OOD might be different.
\end{remark}

\begin{problem} (\emph{\textbf{OOD Detection}} \cite{yang2021generalized, fang2022out}). For a classifier $f$ trained on the training data which is drawn \emph{i.i.d.} from an InD joint distribution $\mathcal{D}_{X_{\text{in}}Y_{\text{in}}}$, given the score function $s$ and a predefined threshold $\delta$, the aim of OOD detection is to reject OOD samples based on a decision function $g: \mathcal{X} \rightarrow \{0,1\}$ such that for any test data $\bm{x}$ drawn from the mixed marginal distribution $\mathcal{D}_{X}$: $g(\bm{x}) = 1$ (inlier, $\bm{x} \in \mathcal{D}_{X_{\text{in}}}$) if $s(\bm{x}) \ge \delta$ and $g(\bm{x}) = 0$ (outlier, $\bm{x} \in \mathcal{D}_{X_{\text{out}}}$) otherwise.
\end{problem}
\begin{definition} (\emph{\textbf{Risk of OOD Detection}}). Given a mixture distribution $p(\bm{x}) = \pi_{\text{in}} p(\bm{x}|\text{in}) + (1-\pi_{\text{in}})p(\bm{x}|\text{out})$, the risk of OOD detection is
\begin{equation}
\label{eq6}
\begin{aligned}
\mathcal{R}_{ood}(g) &= \pi_{\text{in}} \mathbb{E}_{\substack{p(\bm{x}|\text{in})}}[\mathbb{I}(g(\bm{x}) = 0)]\\  &+ (1-\pi_{\text{in}}) \mathbb{E}_{\substack{p(\bm{x}|\text{out})}}[\mathbb{I}(g(\bm{x}) = 1)].
\end{aligned}
\end{equation}
\end{definition}
The above definition indicates that the risk of OOD detection arises from rejecting inlier (seen classes) samples as outlier (unseen classes), and accepting outlier samples as inlier. It is worth mentioning that the risk in Eq.~(\ref{eq6}) can not be optimized directly because a common assumption is that OOD samples only encounter in inference stage and be unavailable in training stage, making OOD detection a challenging task.

Common metrics for OOD detection are the false positive rate at $95\%$ true positive rate (FPR95), the area under the receiver operating characteristic curve (AUROC) and the area under the precision-recall curve (AUPR) \cite{hendrycks2017baseline}. For example, AUROC is a threshold-independent metric reflecting the ranking performance between inlier and outlier samples.
\begin{table*}[!t]
	\caption{Failure prediction performance on CIFAR-10 and CIFAR-100. AURC and E-AURC values are multiplied by $10^3$ for clarity, and all remaining values are percentage. The means and standard deviations over three runs are reported.}
	\vskip -0.1in
	\label{table-1}
	\begin{center}
		\renewcommand\tabcolsep{2pt}
		\newcommand{\tabincell}[2]{\begin{tabular}{@{}#1@{}}#2\end{tabular}}
		\scalebox{0.885}{
			\renewcommand{\arraystretch}{1.05}
			\begin{tabular}{llccccccccccccccc}
				\toprule[1.3pt]
				\multirow{2}*{\tabincell{c}{\textbf{Network}}} & \multirow{2}*{\tabincell{c}{\textbf{Method}}}	&\multicolumn{7}{c}{\textbf{CIFAR-10}} &\multicolumn{7}{c}{\textbf{CIFAR-100}} \\
				\cmidrule{3-9} \cmidrule{11-17}
				&&\textbf{AURC} & \textbf{E-AURC}&
				\textbf{FPR95} &
				\textbf{AUROC} & \textbf{AUPR-S} & \textbf{AUPR-E} & \textbf{ACC} &&\textbf{AURC} & \textbf{E-AURC}&
				\textbf{FPR95} &
				\textbf{AUROC} & \textbf{AUPR-S} & \textbf{AUPR-E} & \textbf{ACC}\\
				\midrule
				\multirow{11}{*}{ResNet110}
				&\cellcolor{mygray}baseline \cite{hendrycks2017baseline} &\cellcolor{mygray}\bftab{9.52} & \cellcolor{mygray}\bftab{7.87} &\cellcolor{mygray}43.33 &\cellcolor{mygray}\bftab{90.13} &\cellcolor{mygray}\bftab{99.17} &\cellcolor{mygray}40.09 &\cellcolor{mygray}94.30 &&\cellcolor{mygray}\bftab{89.05} &\cellcolor{mygray}\bftab{49.71} &\cellcolor{mygray}65.65 &\cellcolor{mygray}\bftab{84.91} &\cellcolor{mygray}\bftab{93.65} &\cellcolor{mygray}\bftab{65.77} &\cellcolor{mygray}73.30 \\
				\cmidrule{2-17}
				& mixup \cite{thulasidasan2019mixup} & 16.27 & 14.84 & \bftab{40.71} & 86.21 & 98.45 & 39.45 & \bftab{94.69} &&87.39 & 53.38 & \bftab{62.95} & 84.60 & 93.27 & 64.38 & \bftab{75.08} \\
				& LS \cite{muller2019does} & 25.89 & 23.96 & 45.98 & 81.41 & 97.48 & 37.89 & 93.86 &&110.60 & 71.98 & 64.62 & 82.40 & 90.79 & 63.97 & 73.53 \\
				& Focal \cite{MukhotiKSGTD20} & 9.89 & 7.98 & 41.40 & 90.66 & 99.16 & \bftab{43.71} & 93.87 &&89.70 & 49.78 & 65.94 & 84.63 & 93.64 & 64.96 & 73.11 \\
				& $L_p$ \cite{YunPLS20} & 13.13 & 11.14 & 43.68 & 88.15 & 98.82 & 42.46 & 93.76 &&122.37 & 74.71 & 66.79 & 82.08 & 90.15 & 65.75 & 70.76 \\
				\cmidrule{2-17}
				& LogitNorm \cite{wei2022logitnorm} & 12.57 & 9.79 & 56.27 & 88.82 & 98.96 & 38.72 & 92.64 && 118.00 & 73.45 & 73.09 & 79.56 & 90.54 & 58.73 & 71.68 \\
				& OE \cite{hendrycks2019deep} & 10.10 & 8.11 & 46.89 & 90.02 & 99.14 & 43.45 & 93.76 &&103.06 & 56.76 & 71.11 & 83.81 & 92.62 & 63.97 & 71.16 \\
				& ODIN \cite{LiangLS18} &20.82 & 19.16 & 59.32 & 79.45 & 97.99 & 28.49 & 94.30 &&167.53 & 127.71 & 79.64 & 68.95 & 84.11 & 46.94 & 73.30 \\
				& energy \cite{liu2020energy} &15.13 & 13.47 & 53.89 & 84.72 & 98.59 & 33.00 & 94.30 &&128.66 & 88.85 & 73.54 & 76.80 & 88.73 & 55.02 & 73.30 \\
				& ReAct \cite{sun2021react} &17.67 & 16.01 & 69.70 & 79.86 & 98.33 & 21.50 & 94.30 &&128.02 & 88.20 & 81.21 & 74.87 & 88.93 & 48.84 & 73.30 \\
				& MLogit \cite{hendrycks2019anomalyseg} &14.93 & 13.27 & 53.01 & 85.00 & 98.61 & 34.15 & 94.30 &&125.38 & 85.57 & 70.61 & 77.73 & 89.12 & 57.16 & 73.30 \\
				
				\midrule
				\multirow{11}{*}{WRN-28-10}
				& \cellcolor{mygray}baseline \cite{hendrycks2017baseline}   &\cellcolor{mygray}\bftab{4.76} &\cellcolor{mygray}\bftab{3.91} &\cellcolor{mygray}30.15 &\cellcolor{mygray}\bftab{93.14} &\cellcolor{mygray}\bftab{99.59} &\cellcolor{mygray}42.03 & \cellcolor{mygray}95.91 &&\cellcolor{mygray}\bftab{46.84} &\cellcolor{mygray}\bftab{27.01} &\cellcolor{mygray}56.64 &\cellcolor{mygray}\bftab{88.50} & \cellcolor{mygray}\bftab{96.79} & \cellcolor{mygray}\bftab{62.85} &\cellcolor{mygray}80.76 \\
				\cmidrule{2-17} 
				& mixup \cite{thulasidasan2019mixup}  & 5.30 & 4.76 & \bftab{29.68} & 90.79 & 99.51 & 38.29 & \bftab{96.71} && 46.91 & 29.84 & \bftab{56.05} & 87.61 & 96.51 & 60.42 & \bftab{82.51} \\
				& LS \cite{muller2019does}  & 17.24 & 16.40 & 34.78 & 83.59 & 98.31 & 39.86 & 95.92 && 55.35 & 36.00 & 57.55 & 86.82 & 95.71 & 61.93 & 80.99 \\
				& Focal \cite{MukhotiKSGTD20}  & 5.65 & 4.83 & 32.94 & 92.20  & 99.50 & 40.01 & 95.98 && 49.34 & 29.21 & 58.63 & 87.70 & 96.53 & 61.65 & 80.62 \\
				& $L_p$ \cite{YunPLS20}  & 6.14 & 5.24 & 33.18 & 91.90 & 99.46 & 42.24 & 95.79 && 57.18 & 36.23 & 57.32 & 86.68 & 95.66 & 62.13 & 80.25 \\
				\cmidrule{2-17} 
				& LogitNorm \cite{wei2022logitnorm} & 5.81 & 4.78 & 46.06 & 91.06 & 99.50 & 35.69 & 95.50 && 72.05 & 48.52 & 66.32 & 82.23 & 94.17 & 55.64 & 79.11 \\
				& OE \cite{hendrycks2019deep} & 4.83 & 3.92 & 38.78 & 93.09 & 99.53 & \bftab{44.10} & 95.53 && 58.05 & 34.29 & 62.96 & 86.36 & 95.88 & 61.39 & 79.01 \\
				& ODIN \cite{LiangLS18} &20.37 & 19.53 & 62.04 & 74.70 & 97.99 & 22.11 & 95.91 &&72.58 & 52.54 & 65.22 & 81.02 & 93.77 & 52.98 & 80.76 \\
				& energy \cite{liu2020energy} &6.91 & 6.06 & 39.13 & 90.47 & 99.37 & 37.02 & 95.91 &&57.30 & 37.27 & 64.15 & 85.05 & 95.58 & 55.75 & 80.76 \\
				& ReAct \cite{sun2021react} &5.36 & 4.52 & 31.60 & 92.16 & 99.53 & 40.77 & 94.78 &&57.02 & 36.98 & 62.33 & 85.43 & 95.60 & 58.06 & 80.76 \\
				& MLogit \cite{hendrycks2019anomalyseg} &6.85 & 6.01 & 37.01 & 90.60 & 99.38 & 38.05 & 95.91 &&56.07 & 36.03 & 61.57 & 85.62 & 95.72 & 58.08 & 80.76 \\
				\midrule	
				\multirow{11}{*}{DenseNet}
				& \cellcolor{mygray}baseline \cite{hendrycks2017baseline}  &\cellcolor{mygray}\bftab{5.66} &\cellcolor{mygray}\bftab{4.27} &\cellcolor{mygray}38.64 &\cellcolor{mygray}\bftab{93.14} &\cellcolor{mygray}\bftab{99.55} &\cellcolor{mygray}41.96 &\cellcolor{mygray}94.78 &&\cellcolor{mygray}66.11 &\cellcolor{mygray}\bftab{37.25} &\cellcolor{mygray}\bftab{62.79} &\cellcolor{mygray}\bftab{86.20} &\cellcolor{mygray}95.43 &\cellcolor{mygray}62.19 &\cellcolor{mygray}76.96 \\
				\cmidrule{2-17} 
				& mixup \cite{thulasidasan2019mixup}  & 9.55 & 8.24 & \bftab{37.21} & 89.87 & 99.14 & 41.42 & \bftab{94.92} && \bftab{63.76} & 37.30 & 63.94 & 86.09 & \bftab{95.61} & 61.78 & \bftab{77.82} \\
				& LS \cite{muller2019does}  & 20.31 & 18.80 & 41.77 & 82.79 & 98.03 & 38.74 & 94.56 && 71.37 & 41.85 & 64.44 & 85.47 & 94.84 & 62.30 & 76.71 \\
				& Focal \cite{MukhotiKSGTD20}  & 6.70 & 5.03 & 37.97 & 92.74  & 99.47 & \bftab{45.99} & 94.29 && 69.83 & 38.69 & 64.98 & 86.07 & 95.22 & 62.44 & 76.11 \\
				& $L_p$ \cite{YunPLS20}  & 13.45 & 11.69 & 41.88 & 87.45 & 98.77 & 41.63 & 94.14 && 81.66 & 46.67 & 66.09 & 84.85 & 94.15 & \bftab{63.04} & 74.74 \\
				\cmidrule{2-17} 
				& LogitNorm \cite{wei2022logitnorm} & 10.89 & 8.78 & 56.59 & 88.70 & 99.07 & 36.45 & 93.59 && 116.35 & 76.48 & 74.81 & 78.14 & 90.32 & 55.59 & 73.13 \\
				& OE \cite{hendrycks2019deep} & 8.23 & 6.48 & 45.86 & 91.44 & 99.33 & 42.51 & 94.14 && 86.96 & 48.29 & 70.39 & 84.25 & 93.90 & 61.91 & 73.51 \\
				& ODIN \cite{LiangLS18} &15.37 & 13.97 & 61.77 & 82.02 & 98.54 & 26.03 & 94.78 &&110.50 & 80.71 & 76.37 & 75.71 & 90.14 & 48.78 & 76.96 \\
				& energy \cite{liu2020energy} &8.60 & 7.20 & 51.31 & 89.21 & 99.25 & 33.81 & 94.78 &&100.13 & 70.34 & 74.46 & 78.03 & 91.38 & 51.02 & 76.96 \\
				& ReAct \cite{sun2021react} &7.80 & 6.40 & 55.06 & 89.93 & 99.33 & 32.06 & 94.78 &&135.73 & 105.94 & 86.93 & 68.05 & 87.28 & 36.69 & 76.96 \\
				& MLogit \cite{hendrycks2019anomalyseg} &8.38 & 6.98 & 48.96 & 89.57 & 99.27 & 35.49 & 94.78 &&96.69 & 66.90 & 70.52 & 79.14 & 91.78 & 53.91 & 76.96 \\
				\bottomrule[1.3pt]					
		\end{tabular}}
	\end{center}
\end{table*}

\setParDis
\noindent
\textbf{Improving OOD detection.} OOD detection has attracted a surge of interest in two directions, \emph{i.e.}, post-hoc and training-time regularization. (1) Some works focus on designing more effective scoring functions for detecting OOD samples, such as ODIN score \cite{LiangLS18}, Mahalanobis distance score \cite{lee2018simple}, Energy score \cite{liu2020energy}, GradNorm score \cite{huang2021importance}, ReAct \cite{sun2021react} and MLogit score \cite{hendrycks2019anomalyseg}. (2) Other methods \cite{hendrycks2019deep, wei2022logitnorm, techapanurak2020hyperparameter, cheng2023average, ma2023towards} address the OOD detection problem by training-time regularization. For example, Hendrycks \emph{et al}. \cite{hendrycks2019deep} leveraged outlier samples to train OOD detectors. Wei \emph{et al}. \cite{wei2022logitnorm} shows logit normalization can mitigate the overconfidence issue for OOD samples.

\setParDef
\subsection{Failure Prediction}
Failure prediction, also known as misclassification detection \cite{hendrycks2017baseline, MoonKSH20} or selective classification \cite{geifman2017selective, el2010foundations}, focuses on distinguishing incorrect ($\mathcal{D}^{\bm{\textcolor{red}{\times}}}_{X_{\text{in}}}$) from correct ($\mathcal{D}^{\textcolor{green}{\checkmark}}_{X_{\text{in}}}$) predictions based on their confidence ranking. Intuitively, if the associated confidence of each misclassified sample is lower than that of any correctly classified samples, we can successfully predict each error made by the classification model. 
\begin{problem} (\emph{\textbf{Failure Prediction}}). For a classifier $f$ trained on the training data  which is drawn \emph{i.i.d.} from an InD joint distribution $\mathcal{D}_{X_{\text{in}}Y_{\text{in}}}$, given the score function $s$ and a predefined threshold $\delta$, the aim of failure prediction is to reject the misclassified InD samples based on the following decision function $g: \mathcal{X} \rightarrow \{0,1\}$ such that for any test data $\bm{x}$ drawn from the InD $\mathcal{D}_{X_{\text{in}}}$: $g(\bm{x}) = 1$ (correct, $\bm{x} \in \mathcal{D}^{\textcolor{green}{\checkmark}}_{X_{\text{in}}}$) if $s(\bm{x}) \ge \delta$ and $g(\bm{x}) = 0$ (misclassified, $\bm{x} \in \mathcal{D}^{\bm{\textcolor{red}{\times}}}_{X_{\text{in}}}$) otherwise.
\end{problem}
\begin{definition} (\emph{\textbf{Risk of Failure Prediction}}). Given a classifier $f$ and a decision function $g$, the risk of failure prediction is
	\begin{equation}
	\label{eq5}
	\begin{aligned}
	\mathcal{R}_{fp}(f,g) &= \mathbb{E}_{\substack{p(\bm{x}|\text{in})}}[c \cdot \mathbb{I}(g(\bm{x}) = 0) \\ & + \mathbb{I}(f(\bm{x}) \neq y) \cdot \mathbb{I}(g(\bm{x}) = 1)],
	\end{aligned}
	\end{equation}
\end{definition}
where $c \in (0,1)$ is the reject cost. Similarly, the risk in Eq.~(\ref{eq5}) can not be optimized directly because DNNs often have near perfect training accuracy and there are few misclassified training samples.

Common metrics for failure prediction include the area under the risk-coverage curve (AURC), the normalized AURC (E-AURC) \cite{GeifmanE17, MoonKSH20}, the FPR95, AUROC, AUPR-Success (AUPR-S) and AUPR-Error (AUPR-E) \cite{hendrycks2017baseline}. Particularly, it has been shown \cite{franc2023optimal} that minimizing the AURC metric is equivalent to minimizing $\mathcal{R}_{fp}(f,g)$.

\setParDis
\noindent
\textbf{Improving failure prediction.} For DNNs, Hendrycks \emph{et al}. \cite{hendrycks2017baseline} firstly established a standard \textbf{baseline} for failure prediction by using MSP. Trust-Score \cite{Jiang2018ToTO} adopts the similarity between the classifier and a nearest-neighbor classifier as a confidence measure. 
%The main drawback of Trust-Score is the lack of practicality and scalability in high-dimensional spaces. 
Some works formulate failure prediction as a supervised binary classification problem. Specifically, ConfidNet \cite{corbiere2021confidence, corbiere2019addressing} and SS \cite{luo2021learning} train auxiliary models to predict confidence by learning the misclassified samples in training set. More recently, Qu \emph{et al}. \cite{qu2023towards} proposed a meta-learning framework that constructs virtual training and testing sets to train an auxiliary model.
However, those methods may fail when the model has a high training accuracy, in which few or even no misclassified examples will exist in training set. CRL \cite{MoonKSH20} improves failure prediction by regularizing the model to learn an ordinal ranking relationship based on the historical correct rate. OpenMix \cite{zhu2023openmix} demonstrates that using the easily available outlier samples coming from non-target classes properly can remarkably help failure prediction. For regression tasks like super-resolution and depth estimation in the field of image enhancement and translation, Upadhyay \emph{et al}. \cite{UpadhyayKCMA22} proposed BayesCap that learns a Bayesian identity mapping
for a frozen model in post-hoc manner.
\setParDef

\section{Does Calibration and OOD Detection Help Failure Prediction?}
In recent years, there has been a surge of research focused on alleviating the overconfidence problem of modern DNNs, and existing methods do help calibration and OOD detection of DNNs. In this section, we empirically investigate the reliability of estimated confidence for failure prediction.
\begin{figure}[t]
	\begin{center}
		\centerline{\includegraphics[width=0.94\columnwidth]{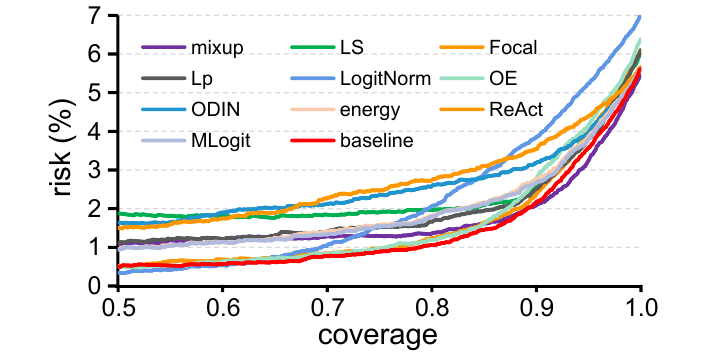}}
		\vskip -0.07 in
		\caption{Comparison of risk-coverage curves. At a given coverage, the lower risk is better. ResNet110 on CIFAR-10.}
		\label{figure-3}
	\end{center}
	\vskip -0.17 in
\end{figure}
\begin{figure*}[!t]
	\begin{center}
		\vskip -0.1 in
		\centerline{\includegraphics[width=0.87\textwidth]{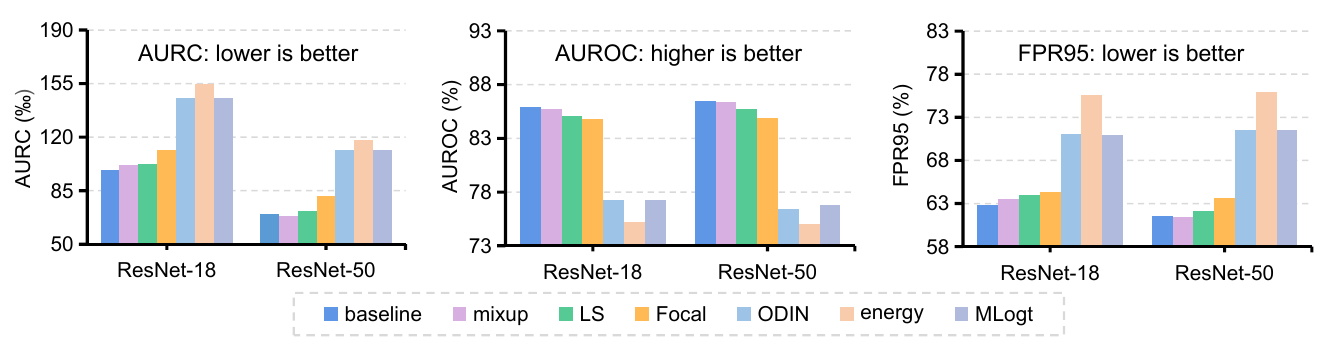}}
		\vskip -0.1 in
		\caption{Large-scale experiments on ImageNet \cite{deng2009imagenet}.}
		\label{figure-4}
	\end{center}
	\vskip -0.15in
\end{figure*}

\subsection{Experimental Setup}
\noindent
\textbf{Datasets and network architectures.} 
We thoroughly conduct experiments on benchmark datasets CIFAR-10 and CIFAR-100 \cite{krizhevsky2009learning}, and large-scale ImageNet \cite{deng2009imagenet} dataset. In terms of network architectures, we consider a range of models: PreAct-ResNet110 \cite{HeZRS16}, WideResNet \cite{zagoruyko2016wide}, DenseNet \cite{HuangLMW17} for experiments on CIFAR-10 and CIFAR-100. For ImageNet, we used ResNet-18 and ResNet-50 \cite{he2016deep} models, respectively. 
%Due to space limitation, we provide the results of more networks like MobileNet \cite{howard2017mobilenets}, EfficientNet \cite{tan2019efficientnet} and more recent architecture ConvMixer \cite{trockman2022patches} in the Supp.Material.

\setParDis
\noindent
\textbf{Evaluation metrics.} We adopt the standard metrics in \cite{hendrycks2017baseline, GeifmanE17, MoonKSH20} to measure failure prediction: AURC, E-AURC, AUROC, FPR95, AUPR-S and AUPR-E. Lower values of AURC, E-AURC, FPR95 and higher values of AUROC, AUPR-S, AUPR-E indicate better failure prediction ability. In addition, we emphasize that test accuracy is also important since we can not sacrifice the original task performance to improve the quality of confidence estimation. Appendix provides detailed definitions of these metrics.

\noindent\textbf{Implementation details.} All models are trained using SGD with a momentum of 0.9, an initial learning rate of 0.1, and a weight decay of 5e-4 for 200 epochs with the mini-batch size of 128 on CIFAR-10 and CIFAR-100. The learning rate is reduced by a factor of 10 at 100, and 150 epochs. We employ a standard data augmentation scheme, \emph{i.e.}, random horizontal flip and $32 \times 32$ random crop after padding with 4 pixels on each side. For each experiment, the mean and standard deviation over three random runs are reported.

\noindent
\textbf{Evaluated confidence estimation methods.} (1) We evaluate various calibration methods including popular training-time regularization like mixup \cite{thulasidasan2019mixup}, label-smoothing (LS) \cite{muller2019does}, focal loss \cite{LinGGHD20}, $L_p$ norm \cite{joo2020revisiting} and post-hoc method TS \cite{guo2017calibration}. Those methods have been verified to be effective in addressing the miscalibration problem of DNNs. (2) We evaluate various OOD detection methods including training-time regularization like LogitNorm \cite{wei2022logitnorm}, outlier exposure (OE) \cite{hendrycks2019deep} and post-hoc methods like ODIN \cite{LiangLS18}, energy \cite{liu2020energy}, ReAct \cite{sun2021react}, MLogit \cite{hendrycks2019anomalyseg}. Many of those methods have recently been proposed and have shown strong OOD detection performance. Supp.Material provides the introduction and hyperparameter setting for each method.
\setParDef

\subsection{Experimental Results} 
In our experiments, we confirmed the positive confidence calibration effects of mixup \cite{thulasidasan2019mixup}, LS \cite{muller2019does}, Focal \cite{LinGGHD20} and $L_p$ norm \cite{joo2020revisiting}. For example, on CIFAR-100, with focal loss, the ECE (\%) can be reduced from 14.98 to 5.66 for ResNet110 and from 8.00 to 1.35 for DenseNet. These observations are consistent with that in \cite{thulasidasan2019mixup, LinGGHD20}. Similarly, in our experiments, the evaluated OOD detection methods \cite{hendrycks2019deep, wei2022logitnorm, liu2020energy, LiangLS18, hendrycks2019anomalyseg} are indeed effective for detecting OOD samples.

\setParDis
\noindent
\textbf{Popular calibration methods can harm failure prediction.}
In practice, users would naturally expect that the calibrated confidence can be used to filter out low-confidence predictions in risk-sensitive applications.
However, if we shift focus to Table~\ref{table-1}, it is evident that those methods generally lead to \textbf{\emph{worse}} failure prediction performance under various metrics. For example, when training with mixup and LS on CIFAR-10/ResNet110, the AURC ($\downarrow$) increases $6.75$ and $16.37$ percentages, respectively. As for post-hoc calibration technique, TS \cite{guo2017calibration} calibrates probabilities by learning a single scalar parameter $T$ for all classes on a hold-out validation set. In our experiments, using validation set and test set to learn the parameter $T$ are denoted as \emph{TS-valid} and \emph{TS-optimal}, respectively. Table~\ref{table-2} shows that TS has negligible improvement for failure prediction.
Those results are counter-intuitive as we expect those methods, which successfully calibrate the confidence, could be useful for failure prediction.
\begin{table}[h]
	\vskip 0.07in
	\caption{Temperature scaling can hardly improve failure prediction. The metric used is AUROC.}
	\vskip -0.03in
	\label{table-2}
	\setlength\tabcolsep{2pt}
	\centering
	\renewcommand{\arraystretch}{1.1}
	\scalebox{0.97}{
		\begin{tabular}{lccccccc}
			\toprule[1.2pt]
			\multirow{2}{*}{\textbf{Method}} & \multicolumn{3}{c}{\textbf{CIFAR-10}} && \multicolumn{3}{c}{\textbf{CIFAR-100}} \\
			\cmidrule(lr){2-4} \cmidrule(lr){6-8}
			&ResNet &WRNet &DenseNet
			 &&ResNet &WRNet &DenseNet
			\\\midrule
			baseline &90.76 &93.24 &92.87 &&85.00 &87.75 &86.46\\
			TS-valid &89.15 &92.30 &91.05 &&84.66 &87.55 &85.82\\
			TS-optimal &90.88 &93.47 &93.11 &&85.01 &87.77 &86.48\\
			\bottomrule[1.2pt]
		\end{tabular}
	}
	\vskip -0.1in
\end{table}

\noindent
\textbf{Popular OOD detection methods can harm failure prediction.}
In practice, users would naturally expect that a good confidence estimator should help filter out both the OOD samples from unknown classes and misclassified InD samples from known classes. 
However, in Table~\ref{table-1}, we observe that OOD detection methods generally lead to \textbf{\emph{worse}} failure prediction performance under various metrics. For example, on CIFAR-100/ResNet110, the recently proposed training-time OOD detection method LogitNorm has $79.56\%$ AUROC ($\uparrow$), which is $5.35\%$ lower than that of baseline. Similarly, those post-hoc OOD detection methods like ODIN, energy, ReAct and MLogit also have negative effect on failure prediction task. This means that those OOD detection methods make it harder to distinguish incorrect from correct predictions based on their confidence ranking.

\noindent
\textbf{Selective risk analysis.} To make intuitive sense of the effect of those methods on failure prediction, Fig.~\ref{figure-3} plots the risk-coverage curve. Specifically, the selective risk is the empirical loss or error rate that trusts the prediction, while coverage is the probability mass of non-rejected predictions \cite{GeifmanE17, MoonKSH20}. Intuitively, a better failure predictor should have low risk at a given coverage. As can be seen from Fig.~\ref{figure-3}, the baseline has the lowest risks compared to other calibration and OOD detection methods, which indicates that using the confidence output by those methods would, unfortunately, increase the risk when making decisions.

\noindent
\textbf{The same observations generalize to large-scale dataset.} Here we verify our observation that those popular calibration and OOD detection methods often harm failure prediction on ImageNet \cite{deng2009imagenet} dataset, which comprises 1000 classes and over 1.2 million images. We train ResNet-18 and ResNet-50 \cite{he2016deep} that achieve $70.20\%$ and $76.14\%$ top-1 classification accuracy, respectively. As shown in Fig.~\ref{figure-4}, similar negative effect can be observed: those widely acknowledged calibration and OOD detection methods yield worse failure prediction performance than baseline.

\setParDef
    
\subsection{Further Discussion and Analysis}
\subsubsection{Discussion on Calibration for Failure Prediction}
\noindent \textbf{Proper scoring rules.} To gain a deeper understanding, we recall the proper scoring rules \cite{gneiting2007strictly}, which is a decades-old concept to evaluate how well the estimated score $S$ explains the observed labels $Y$. The most widely used scoring rules is log-loss: $\phi^{LL}(S,Y):= -\sum_{k=1}^{K}Y_k \text{log}S_k$. Note
that the scoring rules apply to a single sample, and for a dataset, the average of scores across all samples is used. The expected score with rule $\phi$ on the estimated score vector $S$ with respect to the class label $Y$ drawn according to the true posterior distribution $Q$, \emph{i.e.}, $Q_k=
\mathbb{P}(Y = e_k|X)$, is given by $s_{\phi}(S, Q) := \mathbb{E}_{Y \sim Q} [\phi(S, Y )]$. Next, we define \emph{divergence} between $S$ and $Q$ as:
\begin{equation}
\label{eq7}
d_{\phi}(S,Q):= s_{\phi}(S, Q) - s_{\phi}(Q, Q).
\end{equation} 
A scoring rule is said \emph{proper} if its divergence is always non-negative, and \emph{strictly proper} if additionally $d_{\phi}(S,Q)=0$ implies $S = Q$. For instance, log-loss is a strictly proper scoring rule \cite{kull2015novel}, while focal loss is not strictly proper \cite{charoenphakdee2021focal}. Then, we present the scoring rule decomposition as follows:
\begin{proposition} (\emph{\textbf{Calibration-Discrimination Decomposition}}). Let $C$ be the jointly calibrated scores \emph{i.e.}, $C_k = \mathbb{P}(Y = e_k|S = s)$ for $k = 1,...,K.$  the divergence of strictly proper
	scoring rules can be decomposed as \cite{kull2015novel, dimitriadis2021stable, perez2022beyond}:
	\begin{equation}
	\label{eq8}
	\begin{aligned}
	&\mathbb{E}[d_{\phi}(S,Y)] = \underbrace{\mathbb{E}[d_{\phi}(S,Q)]}_{\text{Epistemic}} + \underbrace{\mathbb{E}[d_{\phi}(Q,Y)]}_{\text{Aleatoric}}\\
	&=\underbrace{\mathbb{E}[d_{\phi}(S,C)]}_{\text{Calibration}} + \underbrace{\mathbb{E}[d_{\phi}(C,Q)]}_{\text{Grouping}} + \underbrace{\mathbb{E}[d_{\phi}(Q,Y)]}_{\text{Aleatoric}},
	\end{aligned}
	\end{equation}
\end{proposition}
where the expectation is taken over $Y \sim Q$ and $X$. The final term, aleatoric loss (also known as irreducible loss), is due to inherent uncertainty that exists when assigning a deterministic label to a sample. The epistemic loss is due to the model not being Bayes-optimal classifier. The epistemic loss can be further decomposed into \emph{calibration} and \emph{grouping} loss. Specifically, \emph{\textbf{calibration}} loss measures the difference between estimated score $S$ and the proportion of positives among instances with the same calibrated score $C$; Grouping loss describes that many instances with the same confidence score $S$ while having different true posterior probabilities $Q$. 
Intuitively, grouping loss captures the \emph{\textbf{discrimination}} ability of confidence score for separating samples. 

\begin{figure}[h]
	%\vskip -0.08 in
	\begin{center}
		\centerline{\includegraphics[width=1.03\columnwidth]{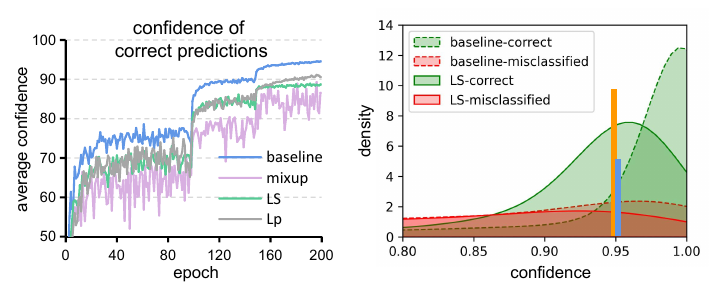}}
		\vskip -0.12 in
		\caption{Average confidence of correct samples during training and confidence distribution of correct and misclassified samples. ResNet110 on CIFAR-10. }
		\label{figure-5}
	\end{center}
	\vskip -0.1 in
\end{figure}

Ideally, the model should be well calibration and with high discrimination.
However, discrimination and calibration do not necessarily keep pace, and many popular methods ignore the discrimination part. Popular calibration methods such as mixup \cite{thulasidasan2019mixup}, LS \cite{muller2019does}, Focal \cite{LinGGHD20} and $L_p$ norm \cite{joo2020revisiting} typically improve calibration by penalizing the confidence of the whole example set to a low level. However, this will lead to undesirable effects: erasing important information about the hardness of samples \cite{shen2020label}, which is useful for keeping the discrimination ability. This is also evident from Fig.~\ref{figure-5}: the average confidence of correctly classified samples is obviously reduced during training, making it harder to separate correctly classified and misclassified examples. Besides, the confidence distribution in Fig.~\ref{figure-5} (Right) also shows that better calibration may lead to worse discrimination: following the definition of grouping loss, we estimate the density of instances with the same confidence score of 0.95, and mark the results in orange (for LS) and blue bars (for baseline). We observe that although LS leads to calibrated confidence, it unfortunately increases the grouping loss (less discrimination). As a result, correct and wrong samples are hard to separate according to the confidence ranking.

Actually, calibration measures an aggregated mismatch between the average accuracy and confidence, without considering the confidence separation between correct and wrong predictions.
%For example, denote $c$ the value of confidence and $w$ the confidence window of a bin, the lowest ECE could achieve when correct and wrong predictions are ``mixed'' in the right way such that those with confidence at level $[c,c+w)$ have a mix of correct and wrong predictions with ratios $c:1-c$. 
While the discrimination alone does not equal to failure prediction because a score estimator with high discrimination may assign high confidence for misclassified samples while low confidence for correctly classified samples. 
Therefore, based on proper score decomposition, good calibration and discrimination are necessary and sufficient characteristics of accurate probabilistic estimation. Section 4 will show that our method can achieve both good failure prediction and calibration performance.

\subsubsection{Discussion on OOD detection for Failure Prediction}
In safety-sensitive applications, both OOD and misclassified InD samples result in significant loss, and therefore should be rejected and handed over to humans. However, as shown in Section 3.2, OOD detection methods often make it harder to detect misclassified samples. To further understand the negative effectiveness of OOD detection methods for failure prediction, we revisit the reject rules of Bayes-optimal classifier for failure prediction and OOD detection, respectively.
\begin{proposition} (\emph{\textbf{Bayes-optimal Reject Rule for Failure Prediction}}). For the risk of failure prediction $\mathcal{R}_{fp}(f,g)$ in Definition~3, the optimal solution $g^*$ of minimizing $\mathcal{R}_{fp}(f,g)$ is given by Chow’s rule  \cite{chow1970optimum, grall2008optimal}:
	\begin{equation}
	\label{profp}
	\begin{aligned}
	g^*(x) = \mathbb{I}({\max}_{y \in \mathcal{Y}}\mathbb{P}(y|\bm{x}) \ge 1-c),
	\end{aligned}
	\end{equation}
\end{proposition}
where $c \in (0,1)$ is the reject cost. This rule can not be directly used because the true posterior probability $\mathbb{P}(y|\bm{x})$ is unknown in practice.
\begin{proposition} (\emph{\textbf{Bayes-optimal Reject Rule for OOD Detection}}). For the risk of OOD detection $\mathcal{R}_{ood}(g)$ in Definition~2, the optimal solution $g^*$ of minimizing $\mathcal{R}_{ood}(g)$ is:
	\begin{equation}
	\label{eq10}
	g^*(x) =  \mathbb{I}([p(\bm{x}|\text{in})/p(\bm{x}|\text{out})] > [(1-\pi_{\text{in}})/\pi_{\text{in}}]),
	\end{equation}
\end{proposition}
where $\pi_{\text{in}} \in (0,1)$ is the mixture ratio of InD and OOD data of the unknown distribution. %We assume that $\mathbb{P}_{\text{out}}$ is unknown in practice, which is consistent with the setting of most existing OOD detection works.
\begin{figure}[t]
	\begin{center}
		\centerline{\includegraphics[width=1.05\columnwidth]{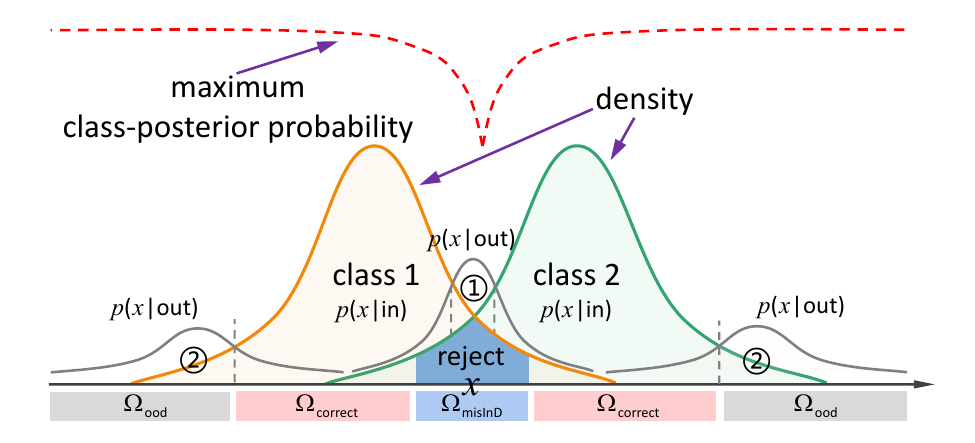}}
		\vskip -0.1 in
		\caption{Illustration of the misalignment between the rejection rules of failure prediction and OOD detection.}
		\label{figure-6}
	\end{center}
	\vskip -0.15 in
\end{figure}

\setParDis
\noindent
\textbf{Misalignment of Reject Regions.} As can be observed from Proposition 2 and Proposition 3, the Bayes-optimal reject rule for failure prediction is based on maximum class-posterior probability ${\max}_{y \in \mathcal{Y}}\mathbb{P}(y|\bm{x})$, while OOD detection rejects samples with small density ratio $p(\bm{x}|\text{in}) / p(\bm{x}|\text{out})$. Fig.~\ref{figure-6} presents an illustration of the reject regions in the case of two Gaussian classes (with the same variance) in $\mathbb{R}$. Specifically, misclassified InD samples are located in the confused region between different classes where the maximum class-posterior probability is low. Based on the Bayes-optimal reject rule, the reject region is $\Omega_{\text{misInD}}$. As for OOD samples, they have no information in the training set and can be noise or semantic-shifted examples from new classes \cite{zhang2023survey, zhu2024openworld, zhu2021prototype, zhu2021class}. The common character of OOD samples is that they are \textbf{\emph{far from the known classes' centroids}} \cite{dubuisson1993statistical}, as shown in region \ding{192} and \ding{193} in Fig.~\ref{figure-6}. Based on the Bayes-optimal reject rule, the reject region $\Omega_{\text{ood}}$ is located on the two sides (\ding{193}) and around the origin (\ding{192}). Particularly, the OOD reject region \ding{192} is overlapped with $\Omega_{\text{misInD}}$. Therefore, the MSP score, which rejects the region between two classes, could serve as a strong, common baseline for both misclassification and OOD detection \cite{hendrycks2017baseline}.\setParDef

\begin{figure}[h]
	\begin{center}
		\vskip -0.1 in
		\centerline{\includegraphics[width=1.06\columnwidth]{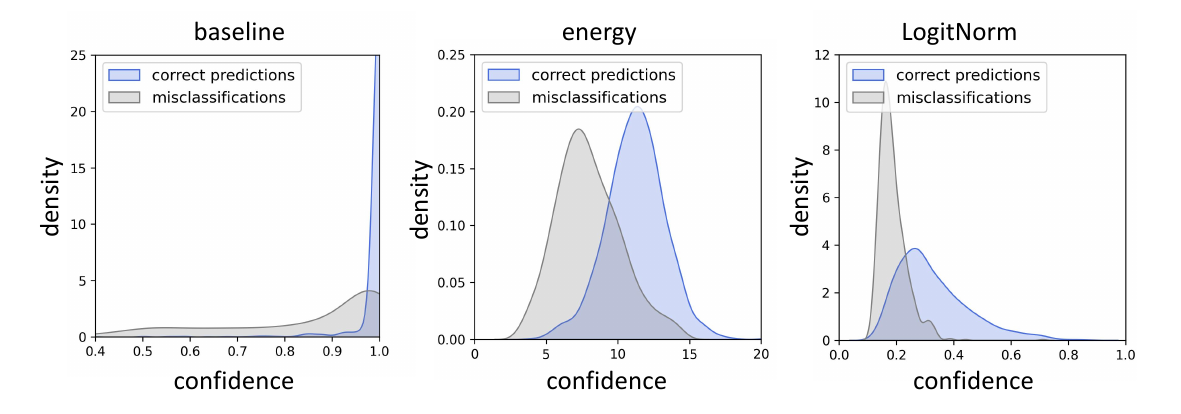}}
		\vskip -0.1 in
		\caption{Due to the misalignment of reject regions, OOD detection methods often lead to worse confidence separation between correct and misclassified InD samples. }
		\label{figure-7}
	\end{center}
	\vskip -0.25 in
\end{figure}

To detect those OOD samples located in \ding{193}, MSP is incompetent because those regions have high MSP score, as shown by the red dotted line in Fig.~\ref{figure-6}. This situation is exactly opposite to the failure prediction. Proposition 3 indicates that the density ratio of InD and OOD is the optimal rule for detecting OOD samples, which can be observed in Fig.~\ref{figure-6}. To this end, many popular OOD detection methods perform density estimation explicitly or implicitly \cite{bitterwolf2022breaking}. For example, Outlier Exposure (OE) \cite{hendrycks2019deep} performs similarly to the binary discrimination between InD and OOD data, and Bayes-optimal solution of Energy-based
OOD detection criterion \cite{liu2020energy} is equivalent to the Bayes-optimal reject rule stated in Proposition 3 \cite{bitterwolf2022breaking}. However, to separate InD and OOD, binary discrimination would compress the confidence distribution of correct and incorrect InD samples, and the density based rule like energy score is unsuitable for detection misclassified samples. Therefore, OOD detection methods often harm failure prediction.   
Fig.~\ref{figure-7} (ResNet110 on CIFAR-10) confirms that those OOD detection methods lead to more overlap between misclassified and correctly classified InD data compared with baseline, \emph{i.e.}, standard training with MSP confidence score. 

\section{Finding Flat Minima for Reliable Confidence Estimation}
As reported in Section 3, none of those popular calibration and OOD detection methods seem to address failure prediction problem (stably) better than simple baseline \cite{hendrycks2017baseline}. In this section, we answer the following two 
fundamental questions: \textbf{(1)} Does there exist a more principled and hassle-free strategy to improve failure prediction? \textbf{(2)} Is it possible that the performance of calibration, OOD detection, and failure prediction be improved concurrently?

\begin{figure*}[t]
	\begin{center}
		%\vskip -0.1 in
		\centerline{\includegraphics[width=0.85\textwidth]{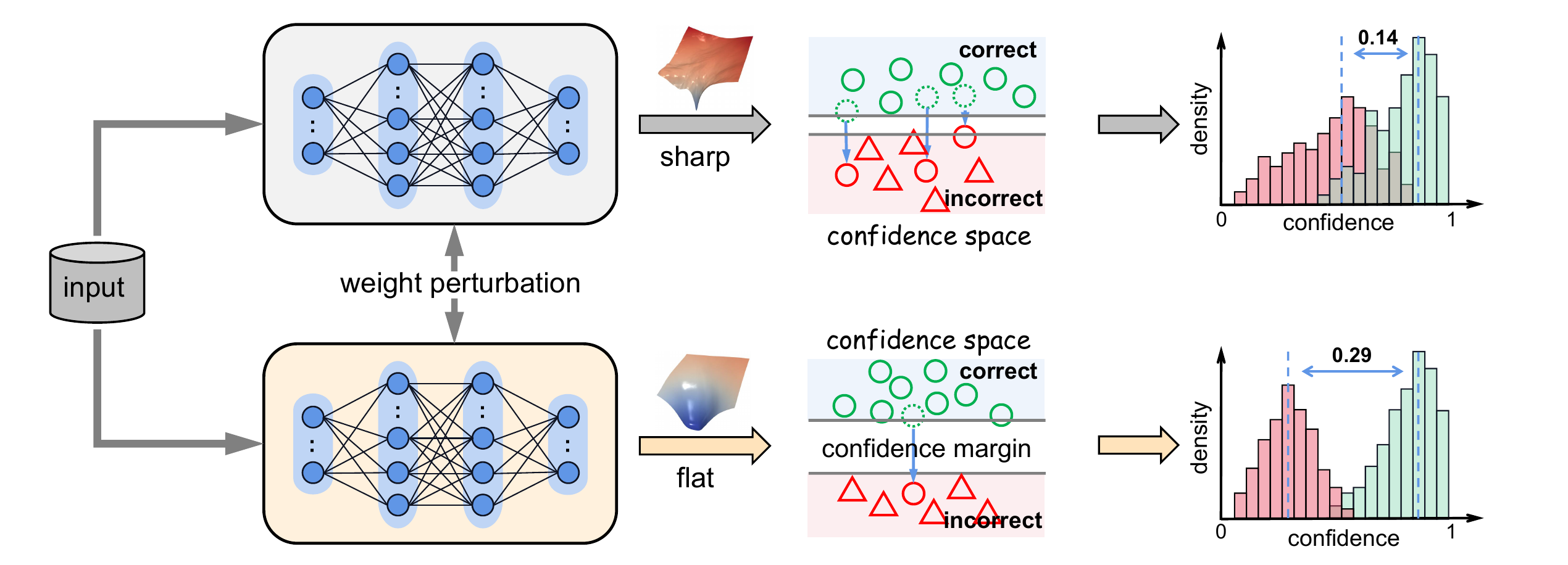}}
		\vskip -0.05 in
		\caption{An intuitive relationship between flatness and failure prediction. The stability of predictions to parameter perturbations can be seen as a \textbf{\emph{confidence margin}} condition. This inspires us to improve failure prediction by finding flat minima.}
		\label{figure-8}
	\end{center}
	\vskip -0.15 in
\end{figure*}
\subsection{Motivation and Methodology}
\subsubsection{Motivation}
\textbf{Connection between flat minima and confidence separation.} Confidence separability between correct and incorrect samples is crucial for failure prediction. Let's consider how confidence separability affects the confidence robustness of correct samples: For a correctly classified sample, to become misclassified, it must reduce the probability on the ground truth class and increase its probability on another (wrong) class. During this process, the confidence margin plays a crucial role: a larger confidence margin could make it harder to change the predicted class label. 
Interestingly, flatness of a model reflects how sensitive the correctly classified samples become misclassified when perturbing the weights of a model \cite{huang2020understanding, izmailov2018averaging, foret2020sharpness}. As illustrated in Fig.~\ref{figure-8}, with flat minima, a correct sample is difficult to misclassify under weight perturbations and vice versa. Therefore, we conjecture that the confidence gap
between correct and incorrect samples of a flat minima is larger than that of a sharp minima.

\setParDis
\noindent
\textbf{Representation learning and Uncertainty.} Misclassification with high confidence means that the sample is mapped into the density region of a wrong class. This is often attributed to spurious correlations appearing in the sample and the wrong class. Flat minima have been theoretically proved to result in invariant and disentangled representations \cite{AchilleS18}, which is effective for spurious representations mitigation \cite{cha2021swad}. Therefore, with fewer spurious or irrelevant representations, the misclassified sample would be near the decision boundary with low confidence and less activated for wrong classes. Besides, it has been shown that flat minima correspond to regions in parameter space with rich posterior uncertainty \cite{pml1Book}. Therefore, flat minima has the advantage of indicating the uncertainty of an input.
\begin{figure}[h]
	\begin{center}
		\vskip -0.15 in
		\centerline{\includegraphics[width=1.02\columnwidth]{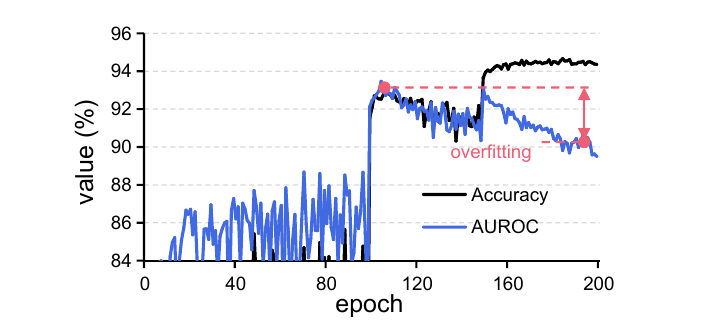}}
		\vskip -0.1 in
		\caption{\emph{Reliable overfitting} phenomenon. Test accuracy continually increases while the AUROC decreases at the last phases. ResNet110 on CIFAR-10.}
		\label{figure-9}
	\end{center}
	\vskip -0.2 in
\end{figure}

\noindent
\textbf{Reliable overfitting phenomenon.} As shown in Fig.~\ref{figure-9}, we observed an interesting phenomenon that the AUROC can be easily overfitting during the training of a model. Concretely, the test accuracy continually increases while the AUROC decreases at the last phases, making it difficult for failure prediction.
We term this phenomenon as ``\textbf{\emph{reliable overfitting}}'', which exists on different model and dataset settings and is somewhat similar to the \emph{robust overfitting} \cite{rice2020overfitting} in adversarial robustness literature \cite{cheng2023adversarial}. Since flat minima has been verified to be effective for alleviating robust overfitting \cite{WuX020, chen2020robust}, we expect it could also benefit failure prediction.\setParDef

\begin{algorithm}[t]
	\caption{FMFP: Flat Minima for Reliable Failure Prediction algorithm}
	\KwIn{Model Weights $\bm{\theta}$, scheduled learning $\alpha$, cycle length $c$, number of iterations $K$, averaging start epoch $S$, neighborhood size $\rho$, the number of past checkpoints to be averaged $s$, loss function $\mathcal{L}$}
	\KwOut{Model trained with FMFP}
	\For{i $\leftarrow$ 1 to K}{ 
		Sample a mini-batch data\\
		Compute gradient $\nabla \mathcal{L(\bm{\theta})}$ of the batch’s training loss\\
		Compute worst-case perturbation $\hat{\bm{\epsilon}} \leftarrow \rho \frac{\nabla \mathcal{L(\bm{\theta})}}{\Vert \nabla \mathcal{L(\bm{\theta})}\Vert_2} $\\
		Gradient update $\bm{\theta} \leftarrow \bm{\theta} - \alpha \nabla \mathcal{L(\bm{\theta} + \hat{\bm{\epsilon}})}$\\
		
		\eIf{$i \ge S$ and \text{mod}$(i, c)=0$} {
			$\bm{\theta}_{\text{FMFP}}^t \leftarrow \frac{{\bm{\theta}_{\text{FMFP}}}^{t-1} \times s + \bm{\theta}^t}{s+1}$
		}{
			$i ++$ \\
		}
	}	
\end{algorithm}

\begin{table*}[!t]
	\caption{Confidence estimation performance on CIFAR-10 and CIFAR-100. AURC and E-AURC values are multiplied by $10^3$ for clarity, and all remaining values are percentage. The means and standard deviations over three runs are reported.}
	\vskip -0.1in
	\label{table-3}
	\begin{center}
		\renewcommand\tabcolsep{4pt}
		\begin{small}
			\newcommand{\tabincell}[2]{\begin{tabular}{@{}#1@{}}#2\end{tabular}}
			\scalebox{0.78}{
				\renewcommand{\arraystretch}{1.03}
				\begin{tabular}{llcccccccccc}
					\toprule[1.3pt]
					\multirow{2}*{\tabincell{c}{\textbf{Network}}} & \multirow{2}*{\tabincell{c}{\textbf{Method}}}	&\multicolumn{10}{c}{\textbf{CIFAR-10}} \\
					\cmidrule{3-12} 
					&&\textbf{AURC}\textbf{$\downarrow$} & \textbf{E-AURC}\textbf{$\downarrow$}&
					\textbf{FPR95}\textbf{$\downarrow$} &
					\textbf{AUROC}\textbf{$\uparrow$} & \textbf{AUPR-S}\textbf{$\uparrow$} & \textbf{AUPR-E}\textbf{$\uparrow$} &\textbf{ACC}\textbf{$\uparrow$} &
					\textbf{ECE}\textbf{$\downarrow$} & \textbf{NLL}\textbf{$\downarrow$}&
					\textbf{Brier}\textbf{$\downarrow$}\\
					\midrule
					\multirow{6}{*}{ResNet110}
					& baseline \cite{hendrycks2017baseline} &9.52$\pm$0.49 & 7.87$\pm$0.49 & 43.33$\pm$0.59 & 90.13$\pm$0.46 & 99.17$\pm$0.05 & 40.09$\pm$1.73 & 94.30$\pm$0.06 & 3.83$\pm$0.10 & 2.72$\pm$0.04 & 9.65$\pm$0.09 \\
					& CRL \cite{MoonKSH20} &6.60$\pm$0.12 & 4.53$\pm$0.07 & 41.00$\pm$0.28 & 93.59$\pm$0.05 & 99.52$\pm$0.01 & 48.29$\pm$1.38 & 93.63$\pm$0.08  & 1.25$\pm$0.08 & 1.90$\pm$0.02 & 9.35$\pm$0.08 \\
					& OpenMix \cite{zhu2023openmix} & 6.31$\pm$0.32 & 4.98$\pm$0.28 & 39.63$\pm$2.36 & 92.09$\pm$0.36 & 99.48$\pm$0.03 & 41.84$\pm$1.42 & \bftab{94.89$\pm$0.20} & 2.72$\pm$0.15 & 2.01$\pm$0.08 & 8.28$\pm$0.30 \\
					\cmidrule{2-12} 
					& SAM & \bftab{4.93$\pm$0.27} & \bftab{3.54$\pm$0.21} & \bftab{33.19$\pm$0.72} & \bftab{94.15$\pm$0.34} & \bftab{99.63$\pm$0.02} & \bftab{47.19$\pm$1.40} & 94.78$\pm$0.19 & 2.55$\pm$0.27 &1.77$\pm$0.09 & 8.33$\pm$0.12\\
					& SWA & 5.79$\pm$0.06 & 4.02$\pm$0.03 & 41.29$\pm$2.16 & 93.77$\pm$0.12 & 99.58$\pm$0.01 & 43.91$\pm$0.63 & 94.10$\pm$0.12 & 1.10$\pm$0.20 & 1.75$\pm$0.02 & 8.78$\pm$0.13 \\
					& \cellcolor{mygray}ours &\cellcolor{mygray}5.33$\pm$0.15 &\cellcolor{mygray}3.71$\pm$0.11 &\cellcolor{mygray}39.37$\pm$0.77 &\cellcolor{mygray}94.07$\pm$0.09 &\cellcolor{mygray}99.61$\pm$0.01 &\cellcolor{mygray}45.71$\pm$0.50 &\cellcolor{mygray}94.36$\pm$0.09 &\cellcolor{mygray}\bftab{0.45$\pm$0.06} &\cellcolor{mygray}\bftab{1.65$\pm$0.02} &\cellcolor{mygray}\bftab{8.28$\pm$0.08} \\
					\midrule
					\multirow{6}{*}{WRN-28-10}
					& baseline \cite{hendrycks2017baseline} &4.76$\pm$0.62 & 3.91$\pm$0.65 & 30.15$\pm$1.98 & 93.14$\pm$0.38 & 99.59$\pm$0.07 & 42.03$\pm$1.45 & 95.91$\pm$0.07 & 2.50$\pm$0.08 & 1.67$\pm$0.01 & 6.71$\pm$0.06 \\
					& CRL \cite{MoonKSH20} &3.99$\pm$0.17 & 2.92$\pm$0.09 & 32.83$\pm$1.17 & 94.37$\pm$0.21 & 99.70$\pm$0.01 & 43.88$\pm$2.13 & 95.42$\pm$0.20 & 0.68$\pm$0.21 & 1.43$\pm$0.01 & 6.88$\pm$0.14 \\
					& OpenMix \cite{zhu2023openmix} & 2.32$\pm$0.15 & 1.91$\pm$0.17 & \bftab{22.08$\pm$1.86} & 94.81$\pm$0.34 & 99.80$\pm$0.02 & 43.62$\pm$4.39 & \bftab{97.16$\pm$0.10} & 1.32$\pm$0.11 & 1.06$\pm$0.02 & 4.51$\pm$0.09 \\
					\cmidrule{2-12} 
					& SAM & 2.97$\pm$0.17 & 2.37$\pm$0.20 & 26.41$\pm$2.82 & 94.52$\pm$0.41 & 99.76$\pm$0.02 & 41.72$\pm$1.64 & 96.54$\pm$0.09 & 1.75$\pm$0.08 & 1.29$\pm$0.03 & 5.57$\pm$0.05 \\
					& SWA & 2.58$\pm$0.08 & 1.88$\pm$0.09 & 25.21$\pm$0.88 & 95.63$\pm$0.17 & 99.81$\pm$0.01 & \bftab{44.84$\pm$1.09} & 96.28$\pm$0.08 & 1.07$\pm$0.11 & 1.14$\pm$0.01 & 5.59$\pm$0.05 \\
					&\cellcolor{mygray}ours &\cellcolor{mygray}\bftab{2.28$\pm$0.03} & \cellcolor{mygray}\bftab{1.67$\pm$0.02} &\cellcolor{mygray}25.20$\pm$1.23 &\cellcolor{mygray}\bftab{95.71$\pm$0.12} &\cellcolor{mygray}\bftab{99.83$\pm$0.00} &\cellcolor{mygray}43.03$\pm$2.16 &\cellcolor{mygray}96.55$\pm$0.08 &\cellcolor{mygray}\bftab{0.57$\pm$0.09} &\cellcolor{mygray}\bftab{1.04$\pm$0.01} & \cellcolor{mygray}\bftab{5.16$\pm$0.04} \\
					
					\midrule
					\multirow{6}{*}{DenseNet}
					& baseline \cite{hendrycks2017baseline} &5.66$\pm$0.45 & 4.27$\pm$0.48 & 38.64$\pm$4.70 & 93.14$\pm$0.65 & 99.55$\pm$0.05 & 41.96$\pm$3.42 & 94.78$\pm$0.16 & 2.76$\pm$0.11 & 1.97$\pm$0.05 & 8.40$\pm$0.16 \\
					& CRL \cite{MoonKSH20} &5.71$\pm$0.24 & 4.07$\pm$0.17 & 39.03$\pm$0.69 & 93.70$\pm$0.14 & 99.57$\pm$0.02 & 45.03$\pm$0.43 & 94.33$\pm$0.13 & 0.64$\pm$0.12 & 1.72$\pm$0.03 & 8.44$\pm$0.14 \\
					& OpenMix \cite{zhu2023openmix} & 4.68$\pm$0.72 & 3.65$\pm$0.69 & 33.57$\pm$3.70 & 93.37$\pm$0.81 & 99.62$\pm$0.07 & 44.45$\pm$3.39 & \bftab{95.51$\pm$0.23} & 1.92$\pm$0.24 & 1.59$\pm$0.02 & 7.03$\pm$0.19 \\
					\cmidrule{2-12} 
					& SAM & 4.19$\pm$0.40 & 3.07$\pm$0.32 & 34.34$\pm$3.65 & 94.28$\pm$0.41 & 99.68$\pm$0.03 & 44.13$\pm$2.11 & 95.30$\pm$0.16 & 1.87$\pm$0.14 & 1.58$\pm$0.07 & 7.24$\pm$0.35 \\
					& SWA & 4.87$\pm$0.09 & 3.39$\pm$0.13 & 36.15$\pm$2.91 & 94.37$\pm$0.30 & 99.65$\pm$0.01 & 46.02$\pm$1.59 & 94.62$\pm$0.10 & 0.95$\pm$0.06 & 1.57$\pm$0.03 & 7.93$\pm$0.15 \\
					&\cellcolor{mygray}ours &\cellcolor{mygray}\bftab{4.09$\pm$0.11} &\cellcolor{mygray}\bftab{2.87$\pm$0.05} &\cellcolor{mygray}\bftab{30.35$\pm$1.72} &\cellcolor{mygray}\bftab{94.82$\pm$0.10} &\cellcolor{mygray}\bftab{99.70$\pm$0.01} &\cellcolor{mygray}\bftab{46.70$\pm$0.26} &\cellcolor{mygray}95.11$\pm$0.16 &\cellcolor{mygray}\bftab{0.51$\pm$0.08} &\cellcolor{mygray}\bftab{1.42$\pm$0.03} &\cellcolor{mygray}\bftab{7.08$\pm$0.21} \\
					\bottomrule
					\toprule
					\multirow{2}*{\tabincell{c}{\textbf{Network}}} & \multirow{2}*{\tabincell{c}{\textbf{Method}}}	&\multicolumn{10}{c}{\textbf{CIFAR-100}} \\
					\cmidrule{3-12} 
					&&\textbf{AURC}\textbf{$\downarrow$} & \textbf{E-AURC}\textbf{$\downarrow$}&
					\textbf{FPR95}\textbf{$\downarrow$} &
					\textbf{AUROC}\textbf{$\uparrow$} & \textbf{AUPR-S}\textbf{$\uparrow$} & \textbf{AUPR-E}\textbf{$\uparrow$} &\textbf{ACC}\textbf{$\uparrow$} &
					\textbf{ECE}\textbf{$\downarrow$} & \textbf{NLL}\textbf{$\downarrow$}&
					\textbf{Brier}\textbf{$\downarrow$}\\
					\midrule
					\multirow{6}{*}{ResNet110}
					& baseline \cite{hendrycks2017baseline} &89.05$\pm$1.39 & 49.71$\pm$1.23 & 65.65$\pm$1.72 & 84.91$\pm$0.13 & 93.65$\pm$0.16 & 65.07$\pm$0.70 & 73.30$\pm$0.25 & 14.98$\pm$0.23 & 13.02$\pm$0.28 & 41.34$\pm$0.50 \\
					& CRL \cite{MoonKSH20} &77.45$\pm$1.07 & 40.30$\pm$1.04 & 65.03$\pm$0.70 & 86.63$\pm$0.28 & 94.91$\pm$0.13 & 66.02$\pm$0.67 & 74.01$\pm$0.10 & 10.13$\pm$0.20 & 10.49$\pm$0.00  & 37.62$\pm$0.04 \\
					& OpenMix \cite{zhu2023openmix} & 73.84$\pm$1.31 & 41.79$\pm$0.77 & 65.22$\pm$1.35 & 85.83$\pm$0.22 & 94.80$\pm$0.10 & 62.87$\pm$1.20 & \bftab{75.77$\pm$0.35} & 11.54$\pm$0.26 & 10.62$\pm$0.13 & 36.54$\pm$0.35 \\
					\cmidrule{2-12} 
					& SAM & 74.09$\pm$0.53 & 40.39$\pm$0.70 & 65.61$\pm$0.19 & 86.08$\pm$0.20 & 94.96$\pm$0.09 & 63.61$\pm$0.23 & 75.19$\pm$0.17 & 10.24$\pm$0.18 & 10.36$\pm$0.13 & 36.71$\pm$0.31 \\
					& SWA & 69.01$\pm$0.96 & 36.51$\pm$0.32 & 64.00$\pm$0.91 & 86.72$\pm$0.26 & 95.48$\pm$0.02 & 64.63$\pm$1.53 & 75.61$\pm$0.40 & 4.40$\pm$0.33  & 8.49$\pm$0.06  & 33.94$\pm$0.34 \\
					&\cellcolor{mygray}ours &\cellcolor{mygray}\bftab{67.08$\pm$1.23} &\cellcolor{mygray}\bftab{34.55$\pm$0.33} &\cellcolor{mygray}\bftab{61.49$\pm$1.47} &\cellcolor{mygray}\bftab{87.38$\pm$0.21} &\cellcolor{mygray}\bftab{95.71$\pm$0.05} &\cellcolor{mygray}\bftab{66.00$\pm$0.97} &\cellcolor{mygray}75.60$\pm$0.39 &\cellcolor{mygray}\bftab{2.30$\pm$0.26}  &\cellcolor{mygray}\bftab{8.28$\pm$0.10}  &\cellcolor{mygray}\bftab{33.30$\pm$0.36} \\
					
					\midrule
					\multirow{6}{*}{WRN-28-10}
					& baseline \cite{hendrycks2017baseline} &46.84$\pm$0.90 & 27.01$\pm$1.23 & 56.64$\pm$1.33 & 88.50$\pm$0.44 & 96.79$\pm$0.14 & 62.85$\pm$1.04 & 80.76$\pm$0.18 & 7.10$\pm$0.20 & 7.95$\pm$0.08 & 28.21$\pm$0.12 \\
					& CRL \cite{MoonKSH20} &42.38$\pm$0.41 & 23.71$\pm$0.18 & 57.09$\pm$2.02 & 89.01$\pm$0.23 & 97.21$\pm$0.01 & 62.09$\pm$1.63 & 81.31$\pm$0.28 & 3.54$\pm$0.20 & 6.95$\pm$0.02 & 26.53$\pm$0.09 \\
					& OpenMix \cite{zhu2023openmix} & 39.61$\pm$0.54 & 23.56$\pm$0.50 & \bftab{55.00$\pm$1.29} & 89.06$\pm$0.11 & 97.25$\pm$0.06 & 62.16$\pm$0.80 & \bftab{82.63$\pm$0.06} & 3.16$\pm$0.50 & 6.71$\pm$0.09 & 24.93$\pm$0.25 \\
					\cmidrule{2-12} 
					& SAM & 42.43$\pm$0.07 & 24.76$\pm$0.28 & 57.50$\pm$1.54 & 88.62$\pm$0.26 & 97.09$\pm$0.03 & 61.62$\pm$1.46 & 81.80$\pm$0.13 & 5.23$\pm$0.09 & 7.15$\pm$0.02 & 26.54$\pm$0.11 \\
					& SWA & 38.10$\pm$0.29 & 20.82$\pm$0.42 & 57.51$\pm$0.42 & 89.69$\pm$0.23 & 97.57$\pm$0.05 & 62.34$\pm$0.97 & 82.00$\pm$0.07 & 7.10$\pm$0.07 & 6.83$\pm$0.04 & 26.32$\pm$0.10 \\
					&\cellcolor{mygray}ours &\cellcolor{mygray}\bftab{36.99$\pm$0.80} &\cellcolor{mygray}\bftab{20.05$\pm$0.61} &\cellcolor{mygray}56.25$\pm$1.54 &\cellcolor{mygray}\bftab{89.91$\pm$0.31} &\cellcolor{mygray}\bftab{97.66$\pm$0.07} &\cellcolor{mygray}\bftab{62.95$\pm$0.76} &\cellcolor{mygray}82.18$\pm$0.20 &\cellcolor{mygray}\bftab{5.84$\pm$0.18} &\cellcolor{mygray}\bftab{6.49$\pm$0.04} &\cellcolor{mygray}\bftab{25.70$\pm$0.18} \\
					\midrule
					
					\multirow{6}{*}{DenseNet}
					& baseline \cite{hendrycks2017baseline} &66.11$\pm$1.56 & 37.25$\pm$1.03 & 62.79$\pm$0.83 & 86.20$\pm$0.04 & 95.43$\pm$0.14 & 62.19$\pm$0.52 & 76.96$\pm$0.20 & 8.00$\pm$0.20 & 9.16$\pm$0.05 & 33.70$\pm$0.16 \\
					& CRL \cite{MoonKSH20} &61.93$\pm$1.97 & 33.89$\pm$0.96 & 62.83$\pm$1.67 & 87.01$\pm$0.13 & 95.86$\pm$0.13 & 63.01$\pm$1.07 & 77.27$\pm$0.40 & 4.76$\pm$0.24 & 8.18$\pm$0.13 & 32.04$\pm$0.48 \\
					& OpenMix \cite{zhu2023openmix} & 53.83$\pm$0.93 & 29.98$\pm$0.20 & 62.22$\pm$1.15 & 87.45$\pm$0.18 & 96.40$\pm$0.04 & 61.42$\pm$1.27 & \bftab{78.97$\pm$0.31} & 4.25$\pm$0.38 & 7.62$\pm$0.15 & 29.92$\pm$0.34 \\
					\cmidrule{2-12} 
					& SAM & 59.51$\pm$0.58 & 32.21$\pm$0.45 & 61.84$\pm$0.79 & 87.44$\pm$0.17 & 96.08$\pm$0.05 & \bftab{63.64$\pm$0.12} & 77.56$\pm$0.15 & 6.09$\pm$0.15 & 8.25$\pm$0.10 & 31.77$\pm$0.28 \\
					& SWA & 56.16$\pm$2.25 & 30.26$\pm$0.70 & 61.96$\pm$1.73 & 87.70$\pm$0.13 & 96.34$\pm$0.11 & 62.87$\pm$1.57 & 78.13$\pm$0.64 & 4.28$\pm$0.42 & 7.50$\pm$0.17 & 30.44$\pm$0.60 \\
					&\cellcolor{mygray}ours &\cellcolor{mygray}\bftab{53.17$\pm$1.70} & \cellcolor{mygray}\bftab{28.93$\pm$0.95} &\cellcolor{mygray}\bftab{61.43$\pm$2.69} & \cellcolor{mygray}\bftab{87.82$\pm$0.22} &\cellcolor{mygray}\bftab{96.53$\pm$0.12} &\cellcolor{mygray}62.95$\pm$0.44 & \cellcolor{mygray}78.58$\pm$0.32 &\cellcolor{mygray}\bftab{2.10$\pm$0.49} &\cellcolor{mygray}\bftab{7.30$\pm$0.11} &\cellcolor{mygray}\bftab{29.75$\pm$0.45} \\
					
					\bottomrule[1.3pt]
			\end{tabular}}
		\end{small}
	\end{center}
\end{table*}

\subsubsection{Methodology}
There are several methods have been proposed to seek flat minima for DNNs \cite{izmailov2018averaging, foret2020sharpness, pittorino2021entropic, chaudhari2019entropy}. We select \emph{stochastic weight averaging} (SWA) \cite{izmailov2018averaging} and \emph{sharpness-aware minimization} (SAM) \cite{foret2020sharpness} as two representative methods due to the simplicity for proofs-of-concept. Specifically, SWA simply averages multiple parameters of the model along the training trajectory as follows:
\begin{equation}
\label{eq10}
\bm{\theta}_{\text{SWA}}^t = \frac{{\bm{\theta}_{\text{SWA}}}^{t-1} \times s + \bm{\theta}^t}{s+1},
\end{equation}
where $s$ denotes the number of past checkpoints to be averaged, $t$ is the training epoch, $\bm{\theta}$ is the current weights, $\bm{\theta}_{\text{SWA}}$ is the averaged weights. While SAM finds the flat minima by directly perturbing the weights. Specifically, denote the batch-size by $m$, and the loss computed on the batch data is $\mathcal{L} = \frac{1}{n}\sum_{j=1}^n \ell_{\text{CE}}((\bm{x}, y); \bm{\theta})$, where $\ell_{\text{CE}}$ is cross-entropy loss. Then, the optimization objective of SAM is:
\begin{equation}
\label{eqsam}
\min\limits_{\bm{\theta}} \mathcal{L} (\bm{\theta} + \bm{\epsilon}^{*}(\bm{\theta})), ~~\text{where}~ \bm{\epsilon}^{*}(\bm{\theta}) \triangleq \mathop{\arg\max}_{||\bm{\epsilon}||_2 \le \rho}  \mathcal{L} (\bm{\theta} + \bm{\epsilon}),
\end{equation}
where $\rho \ge 0$ is a given neighborhood. In practice, to efficiently seek the worst-case perturbation, Eq.~(\ref{eqsam}) can be approximated
by a first-order Taylor expansion of $\mathcal{L} (\bm{\theta} + \bm{\epsilon}^{*}(\theta))$ w.r.t. $\bm{\theta}$ around 0 as follows:
\begin{equation}
\small
\begin{aligned}
\bm{\epsilon}^{*}(\bm{\theta}) &\triangleq \mathop{\arg\max}_{||\bm{\epsilon}||_2 \le \rho}  \mathcal{L} (\bm{\theta} + \bm{\epsilon}) \thickapprox \mathop{\arg\max}_{||\bm{\epsilon}||_2 \le \rho} (\mathcal{L} (\bm{\theta}) + \bm{\epsilon}^\top \nabla_{\bm{\theta}}\mathcal{L} (\bm{\theta})) \\
&= \mathop{\arg\max}_{||\bm{\epsilon}||_2 \le \rho} \bm{\epsilon}^\top \nabla_{\bm{\theta}}\mathcal{L} (\bm{\theta}) \thickapprox \rho \frac{\nabla_{\bm{\theta}} \mathcal{L}(\bm{\theta})}{\Vert \nabla_{\bm{\theta}} \mathcal{L}(\bm{\theta})\Vert}.
\end{aligned}
\label{eq-fmfp}
\end{equation}

Although the SWA and SAM find flat minima based on different mechanisms, we find they both improve the failure prediction performance. This also motivates us to combine them to get better performance. We refer to the combination of them as \textbf{FMFP} (\emph{\textbf{F}lat \textbf{M}inima for \textbf{F}ailure \textbf{P}rediction}). Algorithm 1 gives pseudo-code for the full FMFP algorithm, which is plug-and-play and can be implemented by a few lines of code.

\begin{table*}[t]
	\caption{Confidence estimation on Tiny-ImageNet dataset.}
	\vskip -0.15in
	\label{table-4}
	\begin{center}
		\renewcommand\tabcolsep{6pt}
		\begin{small}
			\newcommand{\tabincell}[2]{\begin{tabular}{@{}#1@{}}#2\end{tabular}}
			\scalebox{0.8}{
				\renewcommand{\arraystretch}{1.2}
				\begin{tabular}{llccccccccc}
					\toprule[1.2pt]
					\textbf{Network} 
					& \textbf{Method}&
					\textbf{AURC}\textbf{$\downarrow$} & \textbf{E-AURC}\textbf{$\downarrow$}&
					\textbf{FPR95}\textbf{$\downarrow$} &
					\textbf{AUROC}\textbf{$\uparrow$} & \textbf{AUPR-S}\textbf{$\uparrow$} & \textbf{AUPR-E}\textbf{$\uparrow$} &\textbf{ACC}\textbf{$\uparrow$} &
					\textbf{ECE}\textbf{$\downarrow$} & \textbf{NLL}\textbf{$\downarrow$}\\
					\midrule
					\multirow{3}{*}{ResNet-18}
					& baseline \cite{hendrycks2017baseline}  & 124.54$\pm$0.97 & 53.13$\pm$0.12 & 62.45$\pm$0.68 & 86.49$\pm$0.11 & 92.54$\pm$0.04 & 75.10$\pm$0.57 &65.92$\pm$0.43 & 10.13$\pm$0.42 & 15.76$\pm$0.14 \\
					& CRL \cite{MoonKSH20}  & 118.05$\pm$1.88 & 49.55$\pm$0.75 & \bftab{60.65$\pm$1.85} & 86.62$\pm$0.37 & 93.10$\pm$0.09 & \bftab{75.47$\pm$1.30} &65.39$\pm$0.51 & 7.56$\pm$0.64 & 14.69$\pm$0.09 \\ 
					& ours  & \bftab{107.01$\pm$1.17} & \bftab{46.88$\pm$0.89} & 62.35$\pm$1.38 & \bftab{86.86$\pm$0.27} & \bftab{93.65$\pm$0.11} & 73.14$\pm$0.75 &\bftab{67.39$\pm$0.07} & \bftab{4.79$\pm$0.20} & \bftab{13.17$\pm$0.05} \\
					
					\midrule
					\multirow{3}{*}{ResNet-50}
					& baseline \cite{hendrycks2017baseline}  & 119.01$\pm$3.19 & 53.05$\pm$1.04 & 63.56$\pm$0.20 & 86.22$\pm$0.20 & 92.66$\pm$0.18 & 73.74$\pm$0.61 &65.96$\pm$0.61 & 10.14$\pm$0.10 & 15.41$\pm$0.26 \\
					& CRL \cite{MoonKSH20}  & 110.80$\pm$2.33 & 48.27$\pm$1.21 & 62.01$\pm$0.35 & 87.02$\pm$0.18 & 93.38$\pm$0.19 & 74.15$\pm$0.26 &66.79$\pm$0.31 & 7.35$\pm$0.29 & 14.23$\pm$0.19 \\ 
					& ours  & \bftab{98.43$\pm$1.17} & \bftab{42.55$\pm$0.89} & \bftab{60.71$\pm$1.38} & \bftab{87.71$\pm$0.27} & \bftab{94.28$\pm$0.11} & \bftab{74.19$\pm$0.75} &\bftab{68.49$\pm$0.10} & \bftab{4.85$\pm$0.20} & \bftab{12.57$\pm$0.05} \\
					\bottomrule[1.3pt]
			\end{tabular}}
		\end{small}
	\end{center}
	%\vskip -0.2in
\end{table*}
\begin{figure*}[t!]
	\begin{center}
		\vskip -0.1 in
		\centerline{\includegraphics[width=0.94\textwidth]{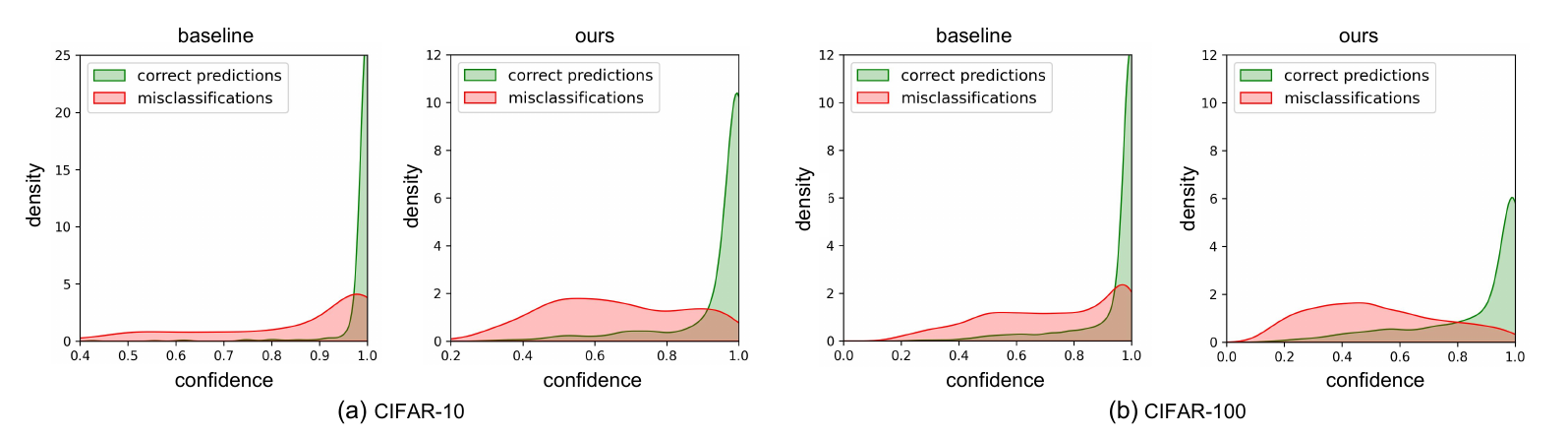}}
		\vskip -0.1 in
		\caption{Our method leads to a better separation between the confidence distribution of correct and misclassified samples, which significantly improves the failure prediction performance.}
		\label{figure-10}
	\end{center}
	\vskip -0.16 in
\end{figure*}

\subsubsection{Theoretical Analysis}
In Section 3.3, we show the Bayes-optimal reject rules for failure prediction and OOD detection, and the misalignment of reject regions. Different from existing works which focus on designing various post-hoc scores (\emph{e.g.}, energy score \cite{liu2020energy}, maxlogit score \cite{hendrycks2019anomalyseg}, ConfidNet score \cite{corbiere2021confidence}), a principled and fundamental approach is to learn a Bayes-like classifier. In the following, we demonstrate that the Bayes classifier prefers flat minima based on the PAC Bayesian framework.

PAC-Bayes \cite{mcallester1998some} is a general framework to understand generalization for many machine learning algorithms. Given a prior distribution $P$ and a posterior distribution $Q$ over the classifier weights $\textbf{w}$, the PAC-Bayes framework bounds the generalization error on the expected error of classifier $f_{\textbf{w}}$ with respect to the KL divergence between the posterior and the prior distributions. Formally, considering a posterior distribution $Q$ of the form $\textbf{w}+\textbf{v}$, where $\textbf{v}$ is a random variable. Then, we have the following theorem:
\begin{theorem} (\emph{\textbf{PAC-Bayesian Bound}} \cite{mcallester1998some, neyshabur2017exploring}). For any $\delta > 0$, with probability at least $1-\delta$ over the draw of $n$ training samples, the expected error of $f_{\textbf{w}+\textbf{v}}$ can be bounded as:
	\begin{equation}
	\label{eq12}
	\begin{aligned}
	\mathbb{E}_{\textbf{v}}[\mathcal{L}(f_{\textbf{w}+\textbf{v}})] &\le \mathbb{E}_{\textbf{v}}[\widehat{\mathcal{L}}(f_{\textbf{w}+\textbf{v}})] \\ &+ 4\sqrt{\frac{1}{n}\biggl[KL(Q||P) + \text{ln}\frac{2n}{\delta}\biggr]},
	\end{aligned}
	\end{equation}
\end{theorem}
where $\mathbb{E}_{\textbf{v}}[\mathcal{L}(f_{\textbf{w}+\textbf{v}})]$ is the expected loss, $\mathbb{E}_{\textbf{v}}[\widehat{\mathcal{L}}(f_{\textbf{w}+\textbf{v}})]$ is the empirical
loss, and their difference yields the generalization error.
Following \cite{neyshabur2017exploring, WuX020}, we choose the perturbation $\textbf{v}$ to be a zero mean spherical Gaussian with variance $\sigma^2$ in every direction, and further set $\sigma = \alpha ||\textbf{w}||$, which makes the second term in Eq.~(\ref{eq12}) become a constant $4\sqrt{\frac{1}{n}\biggl(\frac{1}{2\alpha} + \text{ln}\frac{2n}{\delta}\biggr)}$. Then, substitute $\mathbb{E}_{\textbf{v}}[\widehat{\mathcal{L}}(f_{\textbf{w}+\textbf{v}})]$ with $\widehat{\mathcal{L}}(f_{\textbf{w}}) + \big(\mathbb{E}_{\textbf{v}}[\widehat{\mathcal{L}}(f_{\textbf{w}+\textbf{v}})] - \widehat{\mathcal{L}}(f_{\textbf{w}})\big)$, the expected error of classifier can be bounded (w.p. $1-\delta$ over the training data) as follows:
\begin{equation}
\label{eq13}
\begin{aligned}
\mathbb{E}_{\textbf{v}}[\mathcal{L}(f_{\textbf{w}+\textbf{v}})] &\le \widehat{\mathcal{L}}(f_{\textbf{w}}) + \big(\mathbb{E}_{\textbf{v}}[\widehat{\mathcal{L}}(f_{\textbf{w}+\textbf{v}})] - \widehat{\mathcal{L}}(f_{\textbf{w}})\big) \\ &+ 4\sqrt{\frac{1}{n}\biggl(\frac{1}{2\alpha} + \text{ln}\frac{2n}{\delta}\biggr)},
\end{aligned}
\end{equation}
where $\big(\mathbb{E}_{\textbf{v}}[\widehat{\mathcal{L}}(f_{\textbf{w}+\textbf{v}})] - \widehat{\mathcal{L}}(f_{\textbf{w}})\big)$, which is exactly the expectation of the flatness of weight loss landscape, bounds the generalization gap. Therefore, flat minima technique optimizes the flatness $\big(\mathbb{E}_{\textbf{v}}[\widehat{\mathcal{L}}(f_{\textbf{w}+\textbf{v}})] - \widehat{\mathcal{L}}(f_{\textbf{w}})\big)$ to control the above PAC-Bayes bound, leading to reduced generalization gap and more Bayes-like classifier. 

Based on Bayes-optimal reject rules, the Bayes-like classifier learned by flat minima can improve both failure prediction and OOD detection. Specifically, both the performance of \textbf{(1)} maximum class-posterior reject rule (\emph{e.g.}, MSP score) and \textbf{(2)} density reject rule (\emph{e.g.}, energy score) can be boosted, which will be verified by our experiments.

\subsection{Experiments}
\textbf{Experimental setup.} We conduct experiments on CIFAR-10, CIFAR-100, and Tiny-ImageNet \cite{yao2015tiny} with various network architectures. For comparison methods, we mainly compare our method with baseline \cite{hendrycks2017baseline} and CRL \cite{MoonKSH20}, which is the state-of-the-art approach for failure prediction that outperforms representative bayesian methods \cite{GalG16, KendallG17}. For CRL, our implementation is based on the official open-sourced code. For SWA and FMFP, the cyclical learning rate schedule is used as suggested in \cite{izmailov2018averaging}.
For experiments on CIFAR-10 and CIFAR-100, checkpoints at the 120-th epoch of the baseline
models are used as the initial point of SWA and FMFP. 

\setParDis
\noindent
\textbf{Flat minima improves failure prediction.} Comparative results are summarized in Table~\ref{table-3} and Table~\ref{table-4}. We observe that flat minima based methods: SAM, SWA, and FMFP (ours) consistently outperform the strong baseline, CRL \cite{MoonKSH20} and the state-of-the-art failure prediction method OpenMix \cite{zhu2023openmix} on various metrics. Particularly, FMFP generally yields the best results. For example, in the case of ResNet110, our method has 3.94\% and 2.47\% higher values of AUROC on CIFAR-10 and CIFAR-100, respectively.
\begin{figure}[!t]
	\begin{center}
		\centerline{\includegraphics[width=\columnwidth]{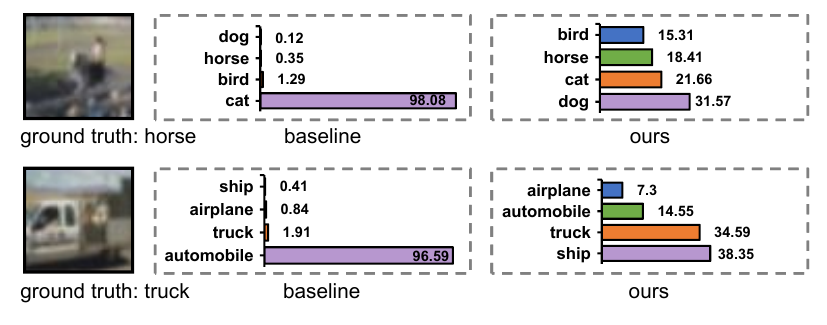}}
		\vskip -0.13 in
		\caption{Predictive distribution on misclassified samples.}
		\label{figure-11}
	\end{center}
	\vskip -0.2 in
\end{figure}

\begin{figure}[!t]
	\begin{center}
		\centerline{\includegraphics[width=1.03\columnwidth]{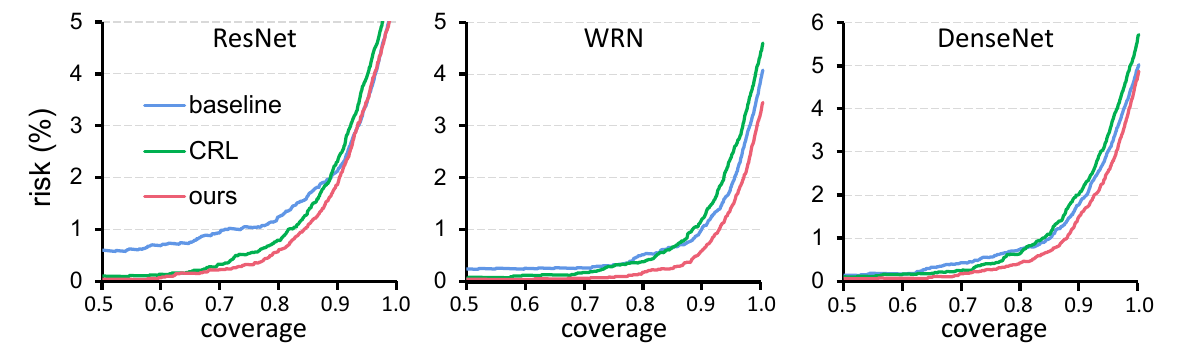}}
		\vskip -0.1 in
		\caption{Comparison of risk-coverage curves on CIFAR-100.}
		\label{figure-12}
	\end{center}
	\vskip -0.2 in
\end{figure}

\noindent
\textbf{Visualization.} In Fig.~\ref{figure-10}, we observe that correct predictions and erroneous predictions of baseline model overlap severely, making it difficult to distinguish them. Our method remarkably shifts the errors’ confidence distributions to smaller values and maintains the confidence of correct samples, which helps to filter out misclassification based on confidence. To be more illustrative, Fig.~\ref{figure-11} presents some examples of misclassified samples on CIFAR-10 and their corresponding confidence distribution. Ours outputs much lower confidence on the erroneously predicted class. Besides, the risk-coverage curves in Fig.~\ref{figure-12} show that our method consistently has the lowest risks at a given coverage. 

\begin{table}[t]
	\vskip 0.05in
	\caption{Comparison with other methods using VGG-16. Results with ``$^{\dagger}$'' and ``$*$'' are from \cite{corbiere2021confidence} and \cite{qu2023towards}, respectively.}
	\vskip -0.12in
	\label{table-5}
	\begin{center}
		\renewcommand\tabcolsep{3pt}
		\begin{small}
			\newcommand{\tabincell}[2]{\begin{tabular}{@{}#1@{}}#2\end{tabular}}
			\scalebox{0.83}{
				\renewcommand{\arraystretch}{1.25}
				\begin{tabular}{lcccc}
					\toprule[1.2pt]
					\multirow{2}*{\tabincell{c}{\textbf{Method}}}	&\multicolumn{4}{c}{\textbf{CIFAR-10}} \\
					\cmidrule{2-5} 
					&\textbf{AURC}\textbf{$\downarrow$} & \textbf{E-AURC}\textbf{$\downarrow$}&
					\textbf{FPR95}\textbf{$\downarrow$}&\textbf{AUROC}\textbf{$\uparrow$}\\
					\midrule
					baseline \cite{hendrycks2017baseline}$^{\dagger}$ &12.66$\pm$0.61 &8.71$\pm$0.50 &49.19$\pm$1.42 &91.18$\pm$0.32\\
					MCDropout \cite{GalG16}$^{\dagger}$ &13.31$\pm$2.63 &9.46$\pm$2.41 &49.67$\pm$2.66 &90.70$\pm$1.96\\
					Trust-Score \cite{Jiang2018ToTO}$^{\dagger}$ &17.97$\pm$0.45 &14.02$\pm$0.34 &54.37$\pm$1.96 &87.87$\pm$0.41 \\
					ConfidNet \cite{corbiere2021confidence}$^{\dagger}$ &11.78$\pm$0.58 &7.88$\pm$0.44 &45.08$\pm$1.58 &92.05$\pm$0.34\\
					SS \cite{luo2021learning}* &-- &-- &44.69 &92.22\\
					DeepEnsembles \cite{lakshminarayanan2017simple}* &-- &-- &45.63$\pm$0.23 & 92.15$\pm$0.03\\
					ConfidNet+Meta \cite{qu2023towards}* &-- &-- &44.56$\pm$0.21 &92.31$\pm$0.02\\ 
					ours &\bftab{6.31$\pm$0.18} &\bftab{4.41$\pm$0.15} &\bftab{38.48$\pm$1.30} &\bftab{93.56$\pm$0.26} \\
					\midrule
					\multirow{2}*{\tabincell{c}{\textbf{Method}}}	&\multicolumn{4}{c}{\textbf{CIFAR-100}} \\
					\cmidrule{2-5} 
					&\textbf{AURC}\textbf{$\downarrow$} & \textbf{E-AURC}\textbf{$\downarrow$}&
					\textbf{FPR95}\textbf{$\downarrow$}&\textbf{AUROC}\textbf{$\uparrow$}\\
					\midrule
					baseline \cite{hendrycks2017baseline}$^{\dagger}$ &113.23$\pm$2.98 &51.93$\pm$1.20 &66.55$\pm$1.56 &85.85$\pm$0.14\\
					MCDropout \cite{GalG16}$^{\dagger}$ &101.41$\pm$3.45 &46.45$\pm$1.91 &63.25$\pm$0.66 &86.71$\pm$0.30\\
					Trust-Score \cite{Jiang2018ToTO}$^{\dagger}$ &119.41$\pm$2.94 &58.10$\pm$1.09 &71.90$\pm$0.93  &84.41$\pm$0.15\\
					ConfidNet \cite{corbiere2021confidence}$^{\dagger}$ &108.46$\pm$2.62 &47.15$\pm$0.95 &62.70$\pm$1.04 &87.17$\pm$0.21\\
					ours &\bftab{73.44$\pm$0.65} &\bftab{36.41$\pm$0.45} &\bftab{61.58$\pm$0.94} &\bftab{87.47$\pm$0.12}\\
					\bottomrule[1.2pt]
			\end{tabular}}
		\end{small}
	\end{center}
	\vskip -0.17in
\end{table}

\noindent
\textbf{Comparison with ConfidNet and its variants.} ConfidNet \cite{corbiere2021confidence, corbiere2019addressing} is a failure prediction method that leverages misclassified samples in training set to train an auxiliary model for confidence estimation. SS \cite{luo2021learning} improves the generalizability of the auxiliary model via steep slope loss.
Qu \emph{et al} \cite{qu2023towards} leveraged a meta-learning framework to train the auxiliary model. The performance of those methods is sensitive to the number of misclassified instances in training set. This issue could be mitigated by leveraging a held-out
validation set to train the auxiliary model, as in \cite{JainLMM23}.
In this paper, we compare those methods on CIFAR-10 and CIFAR-100 with VGG-16 \cite{SimonyanZ14a} following the setting in those works \cite{corbiere2019addressing, luo2021learning, qu2023towards}. As shown in Table~\ref{table-5}, our method consistently outperforms ConfidNet, MCDropout \cite{GalG16} and Trust-Score \cite{Jiang2018ToTO}. Moreover, compared with ConfidNet which is a two-stage training method and MCDropout which involves sampling many (\emph{e.g.}, 100) times to obtain Bayesian inference uncertainty, our method is much simpler and more efficient. 

\begin{figure}[t]
	\begin{center}
		\vskip -0.05 in
		\centerline{\includegraphics[width=\columnwidth]{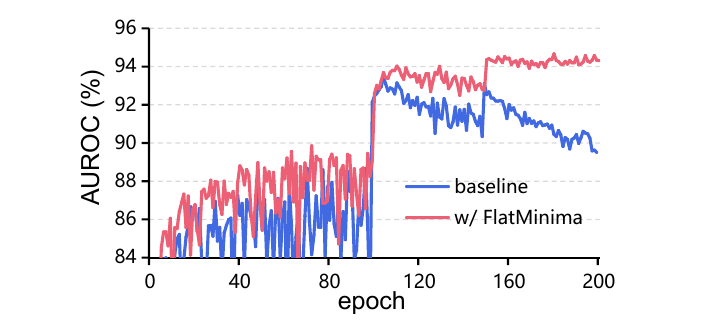}}
		\vskip -0.1 in
		\caption{Flat minima can effectively mitigate the reliable
			overfitting problem, and the AUROC curves continually improved until the end. ResNet110 on CIFAR-10.}
		\label{figure-13}
	\end{center}
	\vskip -0.25 in
\end{figure}

\noindent
\textbf{Flat minima mitigates the reliable overfitting.} Fig.~\ref{figure-13} plots the AUROC curves during training. 
Although the failure prediction performance can be improved by early stopping, the classification accuracy of early checkpoints are much lower (Fig.~\ref{figure-9}). We can clearly observe that with flat minima, reliable overfitting has been diminished significantly, and the AUROC curves robustly improved until the end. Besides, flat minima can further lead to better classification accuracy, avoiding the trade-off between failure prediction and classification accuracy when applying early stopping.

\begin{table}[h]
	\vskip 0.05in
	\caption{Comparison of weights averaging strategy for obtaining flat minima. The model is ResNet110.}
	\vskip -0.12in
	\label{table-6}
	\begin{center}
		\renewcommand\tabcolsep{3pt}
		\begin{small}
			\newcommand{\tabincell}[2]{\begin{tabular}{@{}#1@{}}#2\end{tabular}}
			\scalebox{0.82}{
				\renewcommand{\arraystretch}{1.25}
				\begin{tabular}{lccccc}
					\toprule[1.2pt]
					\multirow{2}*{\tabincell{c}{\textbf{Method}}}	&\multicolumn{5}{c}{\textbf{CIFAR-10}} \\
					\cmidrule{2-6} 
					&\textbf{AURC}\textbf{$\downarrow$} & \textbf{E-AURC}\textbf{$\downarrow$}&
					\textbf{FPR95}\textbf{$\downarrow$}&\textbf{AUROC}\textbf{$\uparrow$}&\textbf{ACC}\textbf{$\uparrow$}\\
					\midrule
					baseline &9.52$\pm$0.49 & 7.87$\pm$0.49 & 43.33$\pm$0.59 & 90.13$\pm$0.46 & 94.30$\pm$0.06\\
					Uniform &5.46$\pm$0.32	&3.92$\pm$0.20	&40.04$\pm$3.44	&93.65$\pm$0.19 &94.52$\pm$0.26
					\\
					Greedy \cite{wortsman2022model} &5.59$\pm$0.31 	&4.15$\pm$0.23 	&41.59$\pm$2.46 &93.14$\pm$0.02 
					&\bftab{94.68$\pm$0.15} \\
					ours &\bftab{5.33$\pm$0.15} &\bftab{3.71$\pm$0.11} &\bftab{39.37$\pm$0.77} &\bftab{94.07$\pm$0.09} &\bftab{94.36$\pm$0.09}\\
					\midrule
					\multirow{2}*{\tabincell{c}{\textbf{Method}}}	&\multicolumn{5}{c}{\textbf{CIFAR-100}} \\
					\cmidrule{2-6} 
					&\textbf{AURC}\textbf{$\downarrow$} & \textbf{E-AURC}\textbf{$\downarrow$}&
					\textbf{FPR95}\textbf{$\downarrow$}&\textbf{AUROC}\textbf{$\uparrow$}&\textbf{ACC}\textbf{$\uparrow$}\\
					\midrule
					baseline &89.05$\pm$1.39 & 49.71$\pm$1.23 & 65.65$\pm$1.72 & 84.91$\pm$0.13 &73.30$\pm$0.25\\
					Uniform &70.32$\pm$0.78 &37.18$\pm$0.46 &63.64$\pm$0.53 &86.74$\pm$0.34 &75.39$\pm$0.36\\
					Greedy \cite{wortsman2022model} &70.54$\pm$0.98 	&37.54$\pm$0.39 	&64.52$\pm$0.30 	&86.62$\pm$0.25 &75.44$\pm$0.33\\
					ours &\bftab{67.08$\pm$1.23} &\bftab{34.55$\pm$0.33} &\bftab{61.49$\pm$1.47} &\bftab{87.38$\pm$0.21} &\bftab{75.60$\pm$0.39}\\
					\bottomrule[1.2pt]
			\end{tabular}}
		\end{small}
	\end{center}
	%\vskip -0.15in
\end{table}

\noindent
\textbf{Comparison of different weights averaging strategies.} Recently, weight averaging has been a popular way to obtain flat minima. For example, Wortsman \emph{et al.} \cite{wortsman2022model} proposed model soups technique that averages the weights of fine-tuned large pre-trained models. Here we verify the effectiveness of \emph{uniform} soup and \emph{greedy} soup \cite{wortsman2022model}. Different from \cite{wortsman2022model} that averages the weights of multiple models fine-tuned with different hyperparameter configurations, we average the model in the later stage of training \emph{i.e.}, the models after 100 epochs are averaged. As reported in Table~\ref{table-6}: \textbf{(1)} Both uniform and greedy weight average strategies are effective for improving confidence estimation. \textbf{(2)} Uniform soup yields slightly worse accuracy but better confidence estimation performance than the greedy strategy. This is reasonable because the greedy strategy emphasizes accuracy, while a model with higher accuracy may suffer from the reliable overfitting problem, as shown in Fig.~\ref{figure-9}. \textbf{(3)} Our method performs better than uniform and greedy strategies.
\setParDef

\begin{table*}[!t]
	\caption{Failure prediction performance in long-tailed recognition.}
	\vskip -0.05in
	\label{table-7}
	\setlength\tabcolsep{4pt}
	\centering
	\renewcommand{\arraystretch}{1.15}
	\scalebox{0.84}{
		\begin{tabular}{lccccccccccccc}
			\toprule[1.2pt]
			\multirow{2}{*}{\textbf{Method}} & \multicolumn{6}{c}{\textbf{CIFAR-10-LT}} & \multicolumn{6}{c}{\textbf{CIFAR-100-LT}} \\
			\cmidrule(lr){2-7} \cmidrule(lr){8-13}
			&\textbf{AURC}\textbf{$\downarrow$} & \textbf{E-AURC}\textbf{$\downarrow$}&
			\textbf{FPR95}\textbf{$\downarrow$} &
			\textbf{AUROC}\textbf{$\uparrow$} &  \textbf{ACC}\textbf{$\uparrow$} &
			\textbf{ECE}\textbf{$\downarrow$} &\textbf{AURC}\textbf{$\downarrow$} & \textbf{E-AURC}\textbf{$\downarrow$}&
			\textbf{FPR95}\textbf{$\downarrow$} &
			\textbf{AUROC}\textbf{$\uparrow$} &  \textbf{ACC}\textbf{$\uparrow$} &
			\textbf{ECE}\textbf{$\downarrow$}  \\\midrule
			LA \cite{MenonJRJVK21} & 62.13$\pm$3.30 & 38.38$\pm$2.00 & 69.77$\pm$0.60 & 84.52$\pm$0.46 & 79.02$\pm$0.63 & 10.21$\pm$0.48 & 347.43$\pm$6.98 & 129.07$\pm$0.86 & 76.47$\pm$1.71 & 78.46$\pm$0.07 & 41.69$\pm$0.70 & 27.93$\pm$0.23\\
			+ CRL \cite{MoonKSH20} & 63.81$\pm$1.55 & 38.83$\pm$0.64 & 63.05$\pm$2.18 & 85.30$\pm$0.43 & 78.50$\pm$0.65 & 5.70$\pm$0.20  & 345.05$\pm$9.21 & 125.64$\pm$3.83 & 76.19$\pm$0.98 & 78.74$\pm$0.58 & 41.58$\pm$0.90 & 27.50$\pm$0.73\\
			+ ours & \bftab{47.78$\pm$0.06} & \bftab{29.33$\pm$0.15} & \bftab{65.73$\pm$2.18} & \bftab{86.41$\pm$0.13} & \bftab{81.42$\pm$0.10} & \bftab{7.09$\pm$0.20}  & \bftab{312.64$\pm$7.49} & \bftab{114.87$\pm$4.89} & \bftab{74.56$\pm$0.74} & \bftab{80.02$\pm$0.59} & \bftab{44.12$\pm$0.34} & \bftab{23.73$\pm$0.21}\\
			\midrule
			CDT \cite{Han01385} & 69.57$\pm$1.25 & 44.55$\pm$0.92  & 71.20$\pm$0.28 & 83.76$\pm$0.27 & 78.48$\pm$0.14 & 17.13$\pm$0.11 & 339.71$\pm$8.46 & 127.40$\pm$0.95 & 79.00$\pm$0.66 & 78.27$\pm$0.28 & 42.40$\pm$0.99 & 34.11$\pm$0.52 \\
			+ CRL \cite{MoonKSH20} & 69.92$\pm$1.86 & 43.94$\pm$1.03 & 70.67$\pm$0.52 & 83.91$\pm$0.16 & 78.09$\pm$0.35 & 17.31$\pm$0.35 & 343.88$\pm$1.43 & 131.31$\pm$4.69 & 78.17$\pm$0.43 & 77.88$\pm$0.63 & 42.36$\pm$0.55 & 33.89$\pm$0.42 \\
			+ ours & \bftab{50.77$\pm$0.69} & \bftab{33.02$\pm$0.85} & \bftab{66.31$\pm$1.13} & \bftab{85.57$\pm$0.50} & \bftab{81.76$\pm$0.21} & \bftab{13.85$\pm$0.09} & \bftab{313.76$\pm$0.76} & \bftab{116.76$\pm$2.77} & \bftab{75.15$\pm$1.32} & \bftab{79.64$\pm$0.43} & \bftab{44.22$\pm$0.28} & \bftab{30.97$\pm$0.31}\\
			\midrule
			VS \cite{KiniPOT21} & 58.45$\pm$2.67 & 37.19$\pm$2.09  & 70.15$\pm$2.08 & 84.47$\pm$0.56 & 80.11$\pm$0.41 & 12.51$\pm$0.23 & 343.48$\pm$6.20 & 129.50$\pm$1.23 & 77.25$\pm$0.63 & 78.20$\pm$0.18 & 42.20$\pm$0.86 & 31.03$\pm$0.33  \\
			+ CRL \cite{MoonKSH20} & 62.06$\pm$0.97 & 39.88$\pm$0.61 & 67.19$\pm$2.05 & 83.98$\pm$0.01 & 79.69$\pm$0.20 & 11.06$\pm$0.64 & 345.06$\pm$7.08 & 128.35$\pm$3.60 & 77.44$\pm$0.81 & 78.29$\pm$0.46 & 41.88$\pm$0.47 & 30.73$\pm$0.81 \\
			+ ours & \bftab{43.34$\pm$1.55} & \bftab{27.93$\pm$1.46} & \bftab{65.60$\pm$1.63} & \bftab{86.21$\pm$0.54} & \bftab{82.97$\pm$0.16} & \bftab{9.66$\pm$0.27}   & \bftab{308.28$\pm$9.88} & \bftab{116.69$\pm$3.48} & \bftab{75.38$\pm$0.29} & \bftab{79.49$\pm$0.40} & \bftab{44.89$\pm$0.93} & \bftab{27.02$\pm$0.90}  \\
			\bottomrule[1.2pt]
		\end{tabular}
	}
	%\vskip -0.1in
\end{table*}

\subsection{Failure Prediction in Long Tailed Recognition}
Existing confidence estimation methods are typically evaluated on balanced training sets. However, the class distributions in real-world settings often follow a long-tailed distribution, in which the head classes have much more training samples than that of the tail classes \cite{cao2019learning, MenonJRJVK21, zhang2023towards}. For example, the class distribution of a disease diagnosis is often long-tailed, \emph{i.e.}, the normal samples are more than the disease samples. In such failure-sensitive applications, reliable confidence estimation is especially crucial.

\setParDis
\noindent\textbf{Experimental setup.} We use two popular long-tailed classification datasets, CIFAR-10-LT and CIFAR-100-LT \cite{cao2019learning}, which are sampled from the original CIFAR dataset over exponential distributions. Following \cite{MenonJRJVK21}, the default imbalance ratio $\rho = 100$ is used and the network is ResNet-32. Our experiments build on code provided by \cite{KiniPOT21}.

\begin{figure}[t]
	\begin{center}
		\vskip -0.05 in
		\centerline{\includegraphics[width=\columnwidth]{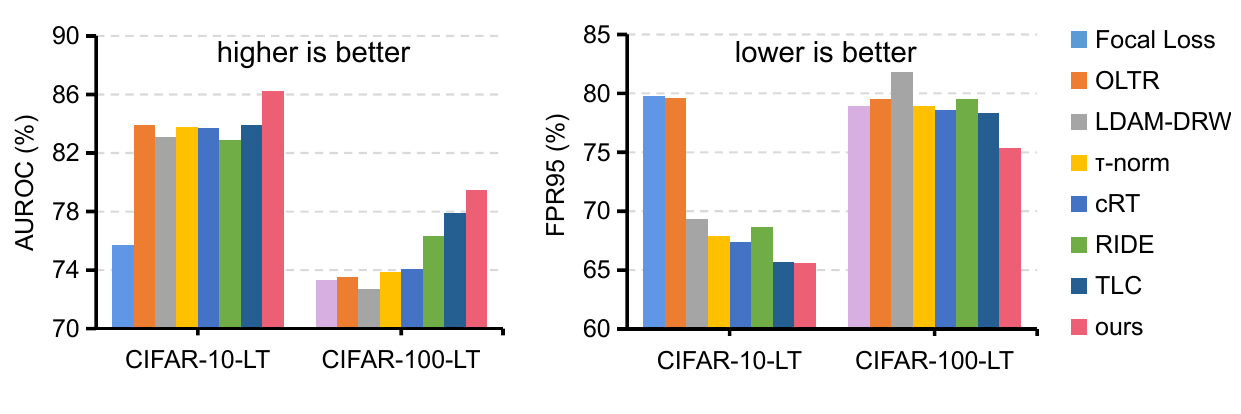}}
		\vskip -0.12 in
		\caption{Comparison with the results in \cite{Li2022CVPR}.}
		\label{figure-14}
	\end{center}
	\vskip -0.3  in
\end{figure}
\noindent\textbf{The challenge of long-tailed failure prediction.} We first examine the performance of CRL on long-tailed setting. Specifically, CRL has been demonstrated to be a strong failure prediction method on balanced datasets. However, as shown in Table~\ref{table-7}, CRL often results in worse performance in both failure prediction and classification accuracy. Moreover, this challenge can not be solved by combining CRL with state-of-the-art long-tailed recognition methods LA \cite{MenonJRJVK21}, CDT \cite{Han01385} and VS \cite{KiniPOT21}. Why does CRL fail on long-tailed recognition? An intuitive explanation is that the tail classes are not well learned by the model, and are thus more likely to result in low confidence predictions like the misclassified samples.

\noindent\textbf{Main results.} As can be observed in Table~\ref{table-7}, our method remarkably improves those long-tailed recognition methods in terms of failure prediction. For example, on CIFAR-10-LT with VS, ours achieves 4.57\% lower average FPR95 ($\downarrow$) and 1.74\% higher average AUROC ($\uparrow$) without degrading the original classification accuracy. Recently, Li \emph{et al.} proposed an evidence-based uncertainty technique named TLC \cite{Li2022CVPR} that outperforms many other methods for long-tailed recognition. In Fig.~\ref{figure-14}, we compare our method (VS based) with TLC under the same experimental setup. The results of TLC and others are from \cite{Li2022CVPR}. As can be observed, our method has consistently better failure prediction performance. \setParDef

\begin{table}[t]
	\vskip 0.05in
	\caption{Failure prediction performance under covariate-shifts. We report the averaged
		results for 15 kinds of corruption under five different levels
		perturbation severity. }
	\vskip -0.03in
	\label{table-8}
	\setlength\tabcolsep{1pt}
	\centering
	\renewcommand{\arraystretch}{1.15}
	\scalebox{0.825}{
		\begin{tabular}{lccccccccc}
			\toprule[1.2pt]
			\multirow{2}{*}{\textbf{Method}} & \multicolumn{3}{c}{\textbf{AUROC}$\uparrow$} & \multicolumn{3}{c}{\textbf{E-AURC}$\downarrow$} & \multicolumn{3}{c}{\textbf{FPR95}$\downarrow$}\\
			\cmidrule(lr){2-4} \cmidrule(lr){5-7} \cmidrule(lr){8-10}
			& ResNet & WRN & DenseNet & ResNet & WRN & DenseNet & ResNet & WRN & DenseNet\\\midrule
			\multicolumn{10}{c}{\textbf{CIFAR-10-C}}  \\
			\midrule
			baseline \cite{hendrycks2017baseline} & 79.92  &83.34 &81.82 &82.61 &63.42 &71.95 &70.23 &64.48 &68.56  \\
			CRL \cite{MoonKSH20} & 82.57  &85.86 &83.86 &68.82 &48.44 &61.87 &68.26 &62.86 &66.93  \\
			ours & \bftab{85.20}  &\bftab{88.10} &\bftab{85.69} &\bftab{52.50} &\bftab{37.15} &\bftab{50.56} &\bftab{64.36} &\bftab{57.38} &\bftab{63.83}  \\
			\bottomrule
			\toprule
			\multicolumn{10}{c}{\textbf{CIFAR-100-C}}  \\ \midrule
			baseline \cite{hendrycks2017baseline} & 77.39  &79.70 &75.86 &128.06 &105.69 &127.13 &76.70 &72.77 &76.88  \\
			CRL \cite{MoonKSH20} & 79.00  &80.71 &78.15 &117.24 &96.67 &112.59 &74.68 &71.13 &75.25  \\
			ours & \bftab{80.56}  &\bftab{82.68} &\bftab{79.29} &\bftab{99.74} &\bftab{82.24} &\bftab{103.51} &\bftab{73.05} &\bftab{69.87} &\bftab{73.59}  \\
			\bottomrule[1.2pt]
		\end{tabular}
	}
\end{table}

\begin{figure}[t]
	\begin{center}
		\centerline{\includegraphics[width=1\columnwidth]{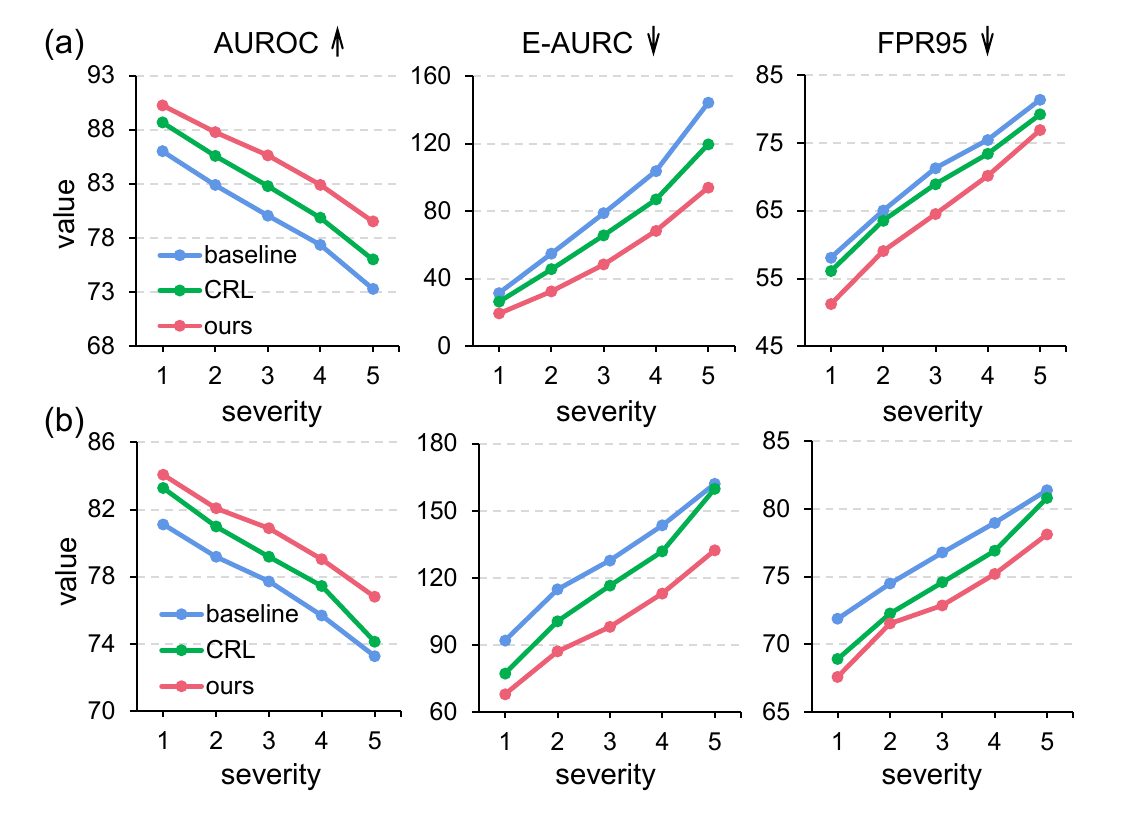}}
		\vskip -0.13 in
		\caption{Failure prediction under covariate-shift: averaged results of 15 types corruptions under different levels of severity on (a) CIFAR-10 and (b) CIFAR-100 with ResNet110.}
		\label{figure-15}
	\end{center}
	\vskip -0.2  in
\end{figure}

\begin{figure*}[t]
	\begin{center}
		\centerline{\includegraphics[width=\textwidth]{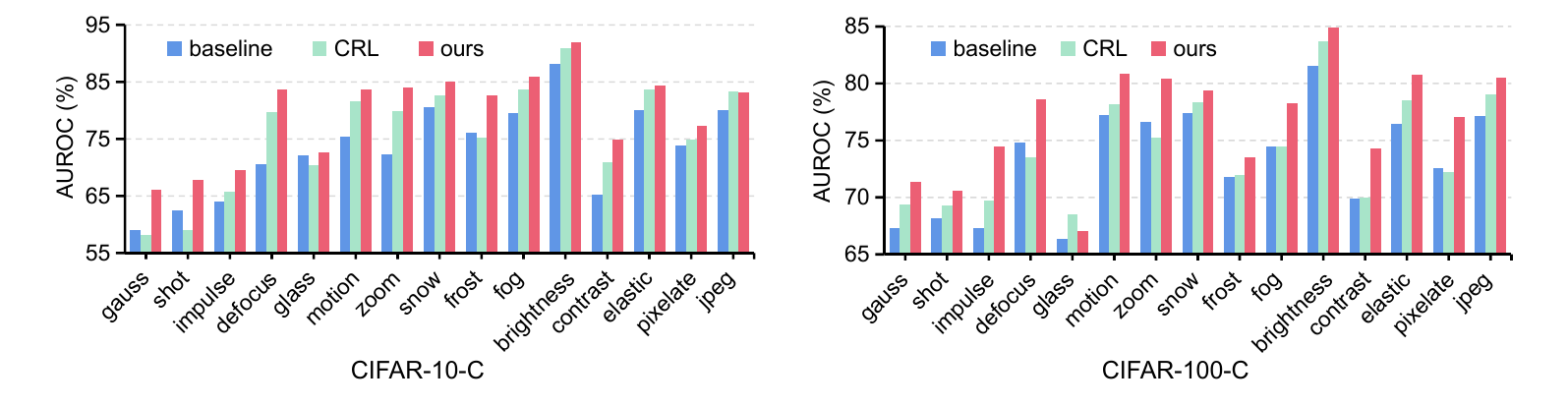}}
		\vskip -0.17 in
		\caption{AUROC (\%) on distribution shift scenarios. CIFAR-10-C and CIFAR-100-C are corruption datasets that contain 15 types of corruption, each with 5 levels of severity. The figure shows the performance on each type of corruption under the severity level of 5. Our method significantly improves the AUROC values of the strong baseline and CRL.}
		\label{figure-16}
	\end{center}
	\vskip -0.2 in
\end{figure*}

\subsection{Failure Prediction under Covariate-Shifts}
Existing confidence estimation works mainly consider the case where the input distribution is static. However, in real-world applications, the deployed systems are running in non-stationary and changing environments and may encode inputs subject to various kinds of covariate-shifts \cite{ovadia2019can}. Take the autonomous driving systems as an example: surrounding environments can be easily changed, \emph{e.g.}, weather change from sunny to cloudy then to rainy \cite{Wang2022CVPR}. The model still needs to make reliable decisions under these domain shift conditions. Therefore, it becomes necessary to evaluate the confidence estimation performance under covariate-shifts. 

\setParDis
\noindent\textbf{Experimental setup.} 
The model is trained on CIFAR-10/100 using the default training setup described in Section 4.2 and evaluated on corrupted dataset CIFAR-10/100-C \cite{HendrycksD19}. Specifically, the corruption dataset contains copies of the original validation set with 15 types of corruptions of algorithmically generated corruptions from noise, blur, weather, and digital categories. Each type of corruption has five levels of severity, resulting in 75 distinct corruptions.

\noindent\textbf{Main results.} Fig.~\ref{figure-15} shows how the averaged results of 15 types corruptions changed when increasing the corruption severity. The average results for all 75 corruptions are reported in Table~\ref{table-8}. As can be seen, our method consistently performs better than baseline and CRL. Fig.~\ref{figure-16} plots the AUROC value of failure prediction on the most severe level-5 corruptions. Firstly, we can observe that all the evaluated methods suffer from a varying degree of performance drop compared with the results on the clean test set. This indicates that the predictions under covariate-shifts are less trustworthy. Secondly, our method outperforms CRL and remarkably improves the performance of baseline. For instance, for the model trained on CIFAR-10, our method has 13.05\% higher values of AUROC under ``defocus'' corruption. \setParDef

\begin{table}[t]
	\vskip 0.05in
	\caption{OOD detection performance. All values are percentages and are averaged over six OOD test datasets.}
	\vskip -0.03in
	\label{table-9}
	\setlength\tabcolsep{1pt}
	\centering
	\renewcommand{\arraystretch}{1.15}
	\scalebox{0.79}{
		\begin{tabular}{lccccccccc}
			\toprule[1.2pt]
			\multirow{2}{*}{\textbf{Method}} & \multicolumn{3}{c}{\textbf{FPR95}$\downarrow$} & \multicolumn{3}{c}{\textbf{AUROC}$\uparrow$} & \multicolumn{3}{c}{\textbf{AUPR}$\uparrow$}\\
			\cmidrule(lr){2-4} \cmidrule(lr){5-7} \cmidrule(lr){8-10}
			& ResNet & WRN & DenseNet & ResNet & WRN & DenseNet & ResNet & WRN & DenseNet\\\midrule
			\multicolumn{10}{c}{\textbf{InD: CIFAR-10}} \\
			\midrule
			baseline \cite{hendrycks2017baseline}  &51.69 &40.83 &48.60 &89.85 &92.32 &91.55 &97.42&97.93 &98.11 \\
			LogitNorm \cite{wei2022logitnorm} &29.72 &12.97 &19.72 &94.29 &97.47 &96.19 &98.70 &99.47 &99.11 \\
			ODIN \cite{LiangLS18} &35.04 &26.94 &30.67 &91.09 &93.35 &93.40 &97.47 &97.98 &98.30 \\
			energy \cite{liu2020energy} &33.98 &25.48 &30.01 &91.15 &93.58 &93.45 &97.49 &98.00 &98.35 \\
			MLogit \cite{hendrycks2019anomalyseg}  &34.61 &26.72 &30.99 &91.13 &93.14 &93.44 &97.46 &97.78 &98.35 \\
			CRL \cite{MoonKSH20} &51.18 &40.83 &47.28 &91.21 &93.67 &92.37 &98.11 &98.67 &98.35 \\
			ours &39.50 &26.83 &35.12 &93.83 &96.22 &94.88 &98.73 &99.23 &98.95 \\
			ours+energy &\bftab{24.71} &\bftab{10.37} &\bftab{15.85} &\bftab{95.75} &\bftab{98.08} &\bftab{96.99} &\bftab{99.10} &\bftab{99.59} &\bftab{99.34} \\
			\bottomrule
			\toprule
			\multicolumn{10}{c}{\textbf{InD: CIFAR-100}}  \\ \midrule
			baseline \cite{hendrycks2017baseline}  &81.68 &77.53 &77.03 &74.21 &77.96 &76.79 &93.34 &94.36 &93.94 \\
			LogitNorm \cite{wei2022logitnorm} &\bftab{63.49} &57.38 &\bftab{61.56} &\bftab{82.50} &86.60 &\bftab{82.10} &\bftab{95.43} &96.80 &\bftab{95.16} \\
			ODIN \cite{LiangLS18} &74.30 &76.03 &69.44 &76.55 &79.57 &80.53 &93.54 &94.59 &94.78 \\
			energy \cite{liu2020energy} &74.42 &74.93 &68.36 &76.43 &79.89 &80.87 &93.59 &94.66 &94.86 \\
			MLogit \cite{hendrycks2019anomalyseg}  &74.45 &75.27 &69.85 &76.61 &79.75 &80.48 &93.66 &94.67 &94.77 \\
			CRL \cite{MoonKSH20} &81.67 &79.08 &75.77 &72.72 &76.81 &76.41 &92.69 &94.22 &93.85 \\
			ours &80.19 &70.98 &79.06 &72.92 &81.54 &71.30 &92.94 &95.71 &91.89 \\
			ours+energy &70.69 &\bftab{54.20} &68.56 &80.22 &\bftab{88.46} &79.81 &94.97 &\bftab{97.38} &94.53 \\
			\bottomrule[1.2pt]
		\end{tabular}
	}
	%\vskip -0.12in
\end{table}

\subsection{Flat Minima Also Improves OOD Detection}
A good confidence estimator should help separate both the OOD and misclassified InD samples from correct predictions. Therefore, besides failure prediction, we explore the OOD detection ability of the proposed flat minima base method.

\setParDis
\noindent\textbf{Experimental setup.} The InD dataset is CIFAR-10 and CIFAR-100. For the OOD datasets, we follow recent works that use six common benchmarks as follows: Textures \cite{CimpoiMKMV14}, SVHN \cite{Netzer2011ReadingDI}, Place365 \cite{ZhouLKO018}, LSUN-C, LSUN-R \cite{YuZSSX15} and iSUN \cite{XuEZFKX15}. During training, the model only sees InD data. At test time, we encounter a mixture of InD and OOD data. For example, the model only sees training set of CIFAR-10 during training; at test time, a mixture of the test set of CIFAR-10 and one OOD dataset. We use the standard metrics in \cite{hendrycks2017baseline} to measure the quality of OOD detection: AUROC, AUPR and FPR95. Supp.Material provides detailed information of the OOD datasets and definitions of the evaluated metrics. 
\begin{figure}[h]
	\begin{center}
		\vskip -0.05in
		\centerline{\includegraphics[width=\columnwidth]{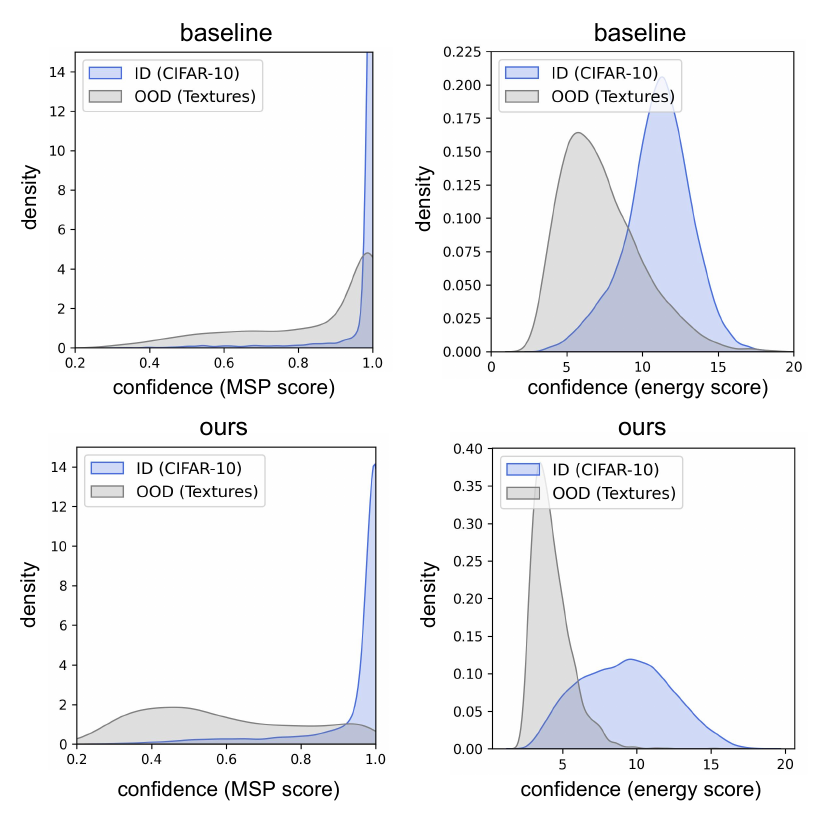}}
		\vskip -0.13 in
		\caption{Confidence distribution of InD and OOD samples.}
		\label{figure-17}
	\end{center}
	\vskip -0.2 in
\end{figure}

\noindent\textbf{Main results.} We report the averaged OOD detection performance over six OOD test datasets in Table~\ref{table-9}. The
results show that our method can achieve state-of-the-art performance on different datasets and networks. In addition, since our flat minima based method is a training-time technique, it can combine with any other post-processing OOD detection methods such as energy and MLogit to get higher OOD detection performance. 
To gain further insights, Fig.~\ref{figure-17} compares the softmax score histogram distributions of baseline and our method on CIFAR-10 with ResNet110. It is obvious that our method makes the scores more distinguishable between InD and OOD, enabling more effective OOD detection.

\setParDef
\section{Concluding Remarks}
Reliable confidence estimation could benefit a wide range of risk-sensitive fields that range from health care (\emph{e.g.,} clinical decision making) to transportation (\emph{e.g.,} autonomous driving), and to business applications. In this paper, with rigorous comprehension and extensive experiments, we rethink the
reliability of both calibration and OOD detection methods in terms of their performances on the failure prediction task. For example, we observe that the simple baseline, \emph{i.e.,} maximum softmax probability score could surprisingly outperform existing methods for detecting classification failures. 
Besides, we extend current evaluation to more realistic setups such as long-tailed and distribution shift scenarios, and propose a unified flat minima based method that yields state-of-the-art confidence estimation performance. We hope to provide machine learning researchers with a deeper understanding of current methods and offer machine learning practitioners a strong baseline that renders safety against classification failures in real-world applications.

\appendices 
\section{Evaluation Metrics}
\subsection{Failure Prediction}
\textbf{AURC \& E-AURC.} AURC measures the area under the
curve drawn by plotting the risk according to coverage. The
coverage indicates the ratio of samples whose confidence
estimates are higher than some confidence threshold, and the
risk, also known as the selective risk \cite{GeifmanE17}, is an error rate computed by using those samples. A low value of AURC implies that correct and incorrect predictions can be well-separable by confidence estimates
associated with samples.
Inherently, AURC is affected by the predictive performance
of a model. To have a unitless performance measure that can
be applied across models, Geifman \emph{et al.}, \cite{GeifmanUE19} introduce
a normalized AURC, named Excess-AURC (E-AURC). Specifically, E-AURC
can be computed by subtracting the optimal AURC,
the lowest possible value for a given model, from the empirical
AURC.

\setParDis\noindent
\textbf{FPR95.} FPR95 can be interpreted as the probability that a negative (misclassified) example is predicted as a correct one when the true positive rate (TPR) is as high as $95\%$. True positive rate can be computed by TPR=TP/(TP+FN), where TP and FN denote the number of true positives and false negatives, respectively. The false positive rate (FPR) can be computed by FPR=FP/(FP+TN), where FP and TN denote the number of false positives and true negatives, respectively.

\noindent
\textbf{AUROC.} 
AUROC measures the area under the receiver operating characteristic curve. The ROC curve depicts the relationship between true positive rate and false positive rate. This metric is a threshold-independent performance evaluation. The AUROC can be interpreted as the probability that a positive example is assigned a higher prediction score than a negative example.

\noindent
\textbf{AUPR-S \& AUPR-E.}
AUPR is the area under the precision-recall curve. The precision-recall curve is a graph showing the precision=TP/(TP+FP) versus recall=TP/(TP+FN). The metrics AUPR-S and AUPR-E indicate the area under the precision-recall curve where correct predictions and errors are specified as positives, respectively.\setParDef

\subsection{Confidence Calibration}
\noindent
\textbf{ECE.} 
Confidence calibration aims to narrow the mismatch between a model's confidence and its accuracy. 
As an approximation of such difference, Expected calibration error (ECE) \cite{Naeini2015ObtainingWC} bins the predictions in $[0,1]$ under $M$ euqally-spaced intervals, and then averages the accuracy/confidence in each bin. Then the ECE can be computed as
\begin{equation}
	\begin{aligned}
		\text{ECE} = \sum_{m=1}^M \frac{\left|B_{m}\right|}{n} \left|\text{acc}(B_{m})-\text{avgConf}(B_{m})\right|,
	\end{aligned}
\end{equation}
where $n$ is the number of all samples. 

\setParDis\noindent
\textbf{NLL.}
Negative log likelihood (NLL) is a standard measure of a probabilistic model's quality \cite{Hastie2001TheEO}, which is defined as
\begin{equation}
	\text{NLL} = -\sum_{i=1}^n \text{log}[\hat{p}(y_{c}|\bm{x_{i}})],
\end{equation}
where $y_{c}$ donates the element for ground-truth class.
In expectation, NLL is minimized if and only if $\hat{p}(Y|\bm{X})$ recovers the truth conditional distribution.

\noindent
\textbf{Brier Score.}
Brier score \cite{Brier1950} can be interpreted
as the average mean squared error between the predicted
probability and one-hot encoded label. It can be computed
as
\begin{equation}
	\text{Brier} = \frac{1}{n}\sum_{i=1}^n\sum_{k=1}^K [\hat{p}(y_{k}|\bm{x_{i}})-t_{k}],
\end{equation}
where $t_{k}=1$ if $k = c$ (ground-truth class), and 0 otherwise.
\setParDef

\subsection{Out-of-distribution Detection}
\noindent
\textbf{AUROC.} AUROC measures the area under the receiver operating characteristic curve, which plots the true positive rate (TPR) of
in-distribution data against the false positive rate (FPR) of
out-of-distribution data by varying a threshold. Thus it can be regarded as an averaged score.

\setParDis
\noindent
\textbf{AUPR-In \& AUPR-Out.} AUPR measures the area under
the precision-recall curve. AUPR-In and AUPR-Out use in and out-of-distribution samples as positives, respectively.

\noindent
\textbf{FPR95.} FPR95 can be interpreted as the probability that OOD example is predicted as ID when the true positive rate (TPR) is as high as $95\%$.
\setParDef

\section{Introduction and hyperparameter setting for each method}
\textbf{Mixup.} Mixup \cite{zhang2018mixup} trains a model on convex combinations of pairs of examples and their labels to encourage linear interpolating predictions. Given a pair of examples $(\bm{x}_{a}, \bm{y}_{a})$ and $(\bm{x}_{b}, \bm{y}_{b})$ sampled from the mini-batch, where $\bm{x}_{a}, \bm{x}_{b}$ represent different samples and $\bm{y}_{a}, \bm{y}_{b}$ denote their one-hot label vectors. Mixup applies linear interpolation to produce augmented data $(\widetilde{\bm{x}}, \widetilde{\bm{y}})$ as follows:
\begin{equation}
	\begin{aligned}
		\widetilde{\bm{x}} = \lambda \bm{x}_{a} + (1-\lambda) \bm{x}_{b}, ~
		\widetilde{\bm{y}} = \lambda \bm{y}_{a} + (1-\lambda) \bm{y}_{b}.
	\end{aligned}
\end{equation}
The $\lambda \in [0,1]$ is a random parameter sampled as $\lambda\sim\text{Beta}(\alpha, \alpha)$ for $\alpha \in (0, \infty)$. Thulasidasan \emph{et~al.} \cite{thulasidasan2019mixup} empirically found that mixup can significantly improve confidence calibration of DNNs. Similar calibration effect of mixup has been verified in natural language processing tasks \cite{mixCalibration}. In our experiments, we follow the setting in \cite{thulasidasan2019mixup} to use $\alpha=0.2$. 

\setParDis\noindent\textbf{Label Smoothing.} Label Smoothing (LS) is commonly used as regularization to reduce overfitting of DNNs. Specifically, when training the model with empirical risk minimization, the one-hot label $\bm{y}$ (\emph{i.e.} the element $y_c$ is 1 for ground-truth class and 0 for others) is smoothed by distributing 
a fraction of mass over the other non ground-truth classes:
\begin{equation}
	\widetilde{\bm{y}_i} = \left\{ 
	\begin{aligned}
		1-\epsilon, &~\text{if} ~~i=c, \\
		\epsilon/(K-1), &~\text{otherwise}.
	\end{aligned}
	\right.
\end{equation}
where $\epsilon$ is a small positive constant coefficient for smoothing the one-hot label, and $K$ is the number of training classes. Recently, Muller \emph{et al}. \cite{muller2019does, thulasidasan2019mixup} showed the favorable calibration effect of LS. It has been shown that an $\epsilon \in [0.05, 0.1]$ performs best for calibration. Therefore, $\epsilon = 0.05$ is used in our experiments.

\noindent\textbf{Focal Loss.} Focal Loss \cite{LinGGHD20} modifies the standard cross entropy loss by weighting loss components of samples in a mini-batch according to how well the model classifies them: 
\begin{equation}
	\mathcal{L}_f := -(1-\hat{p}_{i,y_i})^\gamma ~\text{log}~\hat{p}_{i,y_i},
\end{equation}
where $\gamma$ is a strength coefficient. Intuitively, with focal loss, the gradients of correctly classified samples are restrained and those of incorrectly classified samples are emphasized. Mukhoti \emph{et~al}. \cite{MukhotiKSGTD20} demonstrated that focal loss can automatically learn well-calibrated models. We set the hyperparameter $\gamma=3$ following the suggestion in \cite{MukhotiKSGTD20}.

\noindent\textbf{$L_p$ Norm.} Recently, Joo \emph{et~al}. \cite{joo2020revisiting} explored the effect of explicit regularization strategies (\emph{e.g.}, $L_p$ norm in the logits space) for calibration. Specifically, the learning objective is:
\begin{equation}
	\begin{aligned}
		\mathcal{L}_{L_P}(\bm{x}, y) := \mathcal{L}_{CE}(\bm{x},y) + \lambda \Vert f(\bm{x})\Vert,
	\end{aligned}
\end{equation}
where $f(\bm{x})$ donates the logit of $\bm{x}$, and $\lambda$ is a strength coefficient. Although being simple, $L_p$ norm (\emph{e.g.}, $L_1$ norm) can provide well-calibrated predictive uncertainty \cite{joo2020revisiting}. We use the $L_1$ norm, which is effective for calibration, as shown in \cite{joo2020revisiting}, and $\lambda=0.01$ is used in our experiments.

\noindent\textbf{LogitNorm.} Wei et al., \cite{wei2022logitnorm} thought that the increasing norm of the logit during training leads to the overconfident output. Therefore, they proposed LogitNorm to enforce a constant vector norm on the logits:
\begin{equation}
	\mathcal{L}_{\text{LogitNorm}} := -~\text{log}~\frac{\exp(f_y/(\tau \Vert\bm{f}\Vert)) }{\sum_{i=1}^{k}\exp(f_i/(\tau \Vert\bm{f}\Vert))},
\end{equation}
where $\tau$ denotes the temperature parameter that modulates the magnitude of the logits. Experiments shown that LogitNorm can produce highly distinguishable confidence scores between ID and OOD data.
In our experiments, we use the official open-sourced code in \cite{wei2022logitnorm}.

\noindent\textbf{OE.} OE \cite{hendrycks2019deep} uses auxiliary outliers to help detect
OOD inputs by assigning low confidence for the outliers. Specifically, the following objective is minimized:
\begin{equation}\label{eq2}
	\mathbb{E}_{\mathcal{D}^{\text{train}}_\text{in}} [\ell_\text{CE}(f(\bm{x}),y)] 
	+ \lambda~ \mathbb{E}_{\mathcal{D}_\text{out}} [\ell_\text{OE}(f(\widetilde{\bm{x}}))],
\end{equation}
where $\lambda > 0$ is a penalty hyper-parameter, and $\ell_\text{OE}$ is defined by
Kullback-Leibler (KL) divergence to the uniform distribution: $\ell_\text{OE}(f(\bm{x})) = \text{KL}(\mathcal{U}(y) \Arrowvert
f(\bm{x}))$, in which $\mathcal{U}(\cdot)$ denotes the uniform distribution. In our experiments, we use the official open-sourced code in \cite{hendrycks2019deep}.

\noindent\textbf{ODIN.} ODIN \cite{LiangLS18} leverages
temperature scaling and input perturbations to separate the
softmax score distributions between ID and OOD samples. In our experiments, we set the temperature scaling value to 1000.

\noindent\textbf{energy.} Liu et al., \cite{liu2020energy} found that energy scores better distinguish ID and OOD samples than the widely used softmax scores. Specifically, energy score is computed as:
\begin{equation}
	E(\bm{x}; f) := -\tau ~\text{log}~\sum_{i=1}^{K}\exp(f_i(\bm{x})/\tau),
\end{equation}
where $\tau$ denotes the temperature parameter and $K$ is the number of classes. In our experiments, we set the temperature scaling $\tau$ to 1.

\noindent\textbf{ReAct.} To detect OOD samples, ReAct \cite{sun2021react} truncates the high activations during test time. By rectifying the activations, the outsized contribution of hidden units on OOD output can be attenuated, resulting in a stronger separability from ID data. In our experiments, we set the rectification percentile to 90.

\noindent\textbf{MLogit.} MLogit \cite{hendrycks2019anomalyseg} uses the maximum logit score as the confidence, which outperform the maximum softmax score for OOD detection.

\noindent\textbf{CRL.} Moon et al., \cite{MoonKSH20} proposed CRL to improve the confidence reliability by reflecting the historical correct rate from the view of training dynamics. The training objective of CRL is as follows:
\begin{equation}
	\mathcal{L}_{\text{CRL}}(\bm{x}_i, \bm{x}_j) := \text{max}(0, -g(c_i, c_j)(\kappa_i, \kappa_j)+|c_i - c_j|) 
\end{equation}
where $c_i \in [0,1]$ is the proportion of correct prediction events of $\bm{x}_i$
over the total number of examinations, and $\kappa_i$ denotes a confidence function (e.g., the maximum class probability, negative entropy, and margin). In our experiments, the implementation is based on the official open-sourced code.

\noindent\textbf{OpenMix.} Inspired by the success of Outlier Exposure \cite{hendrycks2019deep} in OOD detection, OpenMix \cite{zhu2023openmix} exploits the easily available outlier
samples, i.e., unlabeled samples coming from non-target
classes, for helping detect misclassification errors. Specifically, it introduces two strategies, named learning with reject class and outlier transformation to improve the failure prediction ability of DNNs.
\setParDef

\section{Experimental details}
For SWA and FMFP, the cyclical learning rate schedule is used as suggested in \cite{izmailov2018averaging}. For experiments on CIFAR-10 and CIFAR-100, checkpoints at 120-th epoch of the baseline models are used as the initial point of SWA and FMFP. For experiments on Tiny-ImageNet \cite{yao2015tiny}, the models (ResNet-18 and ResNet-50) are trained from sketch using SGD with a momentum of 0.9, an initial learning rate of 0.1, and a weight decay of 5e-4 for 90 epochs with the mini-batch size of 128. The learning rate is reduced by a factor of 10 at 40, and 70 epochs. Checkpoints at 50-th epoch of the baseline models are used as the initial point of SWA and FMFP. 
The details of the OOD datasets used for OOD detection are as follows.
\begin{itemize} 
	\item Textures \cite{CimpoiMKMV14} is an evolving collection of textural images in the wild, consisting of 5640 images, organized into 47
	items.
	\item SVHN \cite{Netzer2011ReadingDI} contains 10 classes comprised of the digits 0-9 in street view, which contains 26,032 images for testing.
	\item Place365 \cite{ZhouLKO018} consists in 1,803,460 large-scale photographs of scenes. Each photograph belongs to one of 365 classes.
	\item LSUN \cite{YuZSSX15} has a test set of 10,000 images of 10 different scene categories. Following existing works, we construct two
	datasets, LSUN-C and LSUN-R, by randomly cropped and downsampling LSUN test set, respectively.
	\item iSUN \cite{XuEZFKX15} is a ground truth of gaze traces on images from the SUN dataset. The dataset contains 2,000 images for the
	test.
\end{itemize}

%\newpage
{
	\bibliographystyle{unsrt}
	\bibliography{reference}
}

\end{document}